\documentclass[preprint,12pt]{elsarticle}

\usepackage{amsmath}  
\usepackage{bm}      
\usepackage{amssymb} 
\usepackage{graphicx} 

\usepackage{xcolor}  

\usepackage{booktabs}  
\usepackage{colortbl}  
\usepackage{multirow}  
\usepackage{float}     
\usepackage{adjustbox} 
\definecolor{RuleGray}{gray}{0.7}
\newcommand{\greyrule}{\arrayrulecolor{RuleGray}\midrule\arrayrulecolor{black}}
\floatstyle{plaintop}
\restylefloat{table}

\usepackage[hidelinks]{hyperref} 

\usepackage{pifont}  

\usepackage{subcaption}  
\usepackage{enumitem}    
\usepackage{tikz}
\usetikzlibrary{positioning, arrows.meta, fit, backgrounds, calc}
\usepackage{algorithm}
\usepackage{algpseudocode}
\algrenewcommand\algorithmicrequire{\textbf{Input:}}
\algrenewcommand\algorithmicensure{\textbf{Output:}}

\newcommand{\bsym}[1]{\boldsymbol{#1}}          
\newcommand{\mat}[1]{\ensuremath{\mathsf{#1}}}  
\newcommand{\hp}[1]{\ensuremath{\hat{#1}}}  

\newcommand\p[2]{\frac{\partial #1}{\partial #2}}

\newcommand\pline[2]{\partial #1 / \partial #2} 

\DeclareMathOperator{\sgn}{sgn}                                          
\newcommand{\condmax}[0]{\kappa_{\max}}                                  

\newcommand{\eg}[0]{e.g.\ }
\newcommand{\ie}[0]{i.e.\ }

\definecolor{Red}{RGB}{235,40,25}

\newcommand{\green}[1]{\textcolor{green!60!black}{#1}}
\newcommand{\red}[1]{\textcolor{red}{#1}}

\newcommand{\one}[0]{\ensuremath{\textbf{1}}}   
\newcommand{\rdot}[0]{\ensuremath{\dot{r}}}     

\newcommand{\avec}[0]{\ensuremath{\bsym{a}}}

\newcommand{\cvec}[0]{\ensuremath{\bsym{c}}}

\newcommand{\fvec}[0]{\ensuremath{\bsym{f}}}

\newcommand{\hvec}[0]{\ensuremath{\bsym{h}}}

\newcommand{\kvec}[0]{\ensuremath{\bsym{k}}}

\newcommand{\qvec}[0]{\ensuremath{\bsym{q}}}
\newcommand{\rvec}[0]{\ensuremath{\bsym{r}}}
\newcommand{\svec}[0]{\ensuremath{\bsym{s}}}

\newcommand{\xvec}[0]{\ensuremath{\bsym{x}}}
\newcommand{\yvec}[0]{\ensuremath{\bsym{y}}}

\newcommand{\alphavec}[0]{\ensuremath{\bsym{\alpha}}}

\newcommand{\gammavec}[0]{\ensuremath{\bsym{\gamma}}}

\newcommand{\Amat}[0]{\mat{A}}
\newcommand{\Bmat}[0]{\mat{B}}
\newcommand{\Cmat}[0]{\mat{C}}

\newcommand{\Gmat}[0]{\mat{G}}
\newcommand{\Hmat}[0]{\mat{H}}
\newcommand{\Idty}[0]{\mat{I}}    

\newcommand{\Kmat}[0]{\mat{K}}    
\newcommand{\Lmat}[0]{\mat{L}}    

\newcommand{\Wmat}[0]{\mat{W}}    
\newcommand{\Xmat}[0]{\mat{X}}    

\newcommand{\kSE}[0]{\ensuremath{k_{\text{SE}}}}
\newcommand{\kM}[0]{\ensuremath{k_{\text{M}{\frac{5}{2}}}}}
\newcommand{\kMthreehalf}[0]{\ensuremath{k_{\text{M}{\frac{3}{2}}}}}
\newcommand{\kRQ}[0]{\ensuremath{k_{\text{RQ}}}}    
\newcommand{\kPer}[0]{\ensuremath{k_{\text{P}}}}    

\newcommand{\hpsigK}{\ensuremath{\hp{\sigma}_{K}}}               

\newcommand{\hpstdfval}{\ensuremath{\hp{\sigma}_f}}            
\newcommand{\hpgamma}{\ensuremath{\hp{\gamma}}}                
\newcommand{\hpa}{\ensuremath{\hp{a}}}                         
\newcommand{\hprho}{\ensuremath{\hp{\rho}}}                      
\newcommand{\hpalpha}{\ensuremath{\hp{\alpha}}}                    
\newcommand{\hpavec}{\ensuremath{\hp{\avec}}}                  
\newcommand{\hpgammavec}{\ensuremath{\hp{\gammavec}}}          

\newcommand{\Hmatgam}[1]{\ensuremath{\Hmat^{(\hpgamma, #1)}}}  
\newcommand{\Hmatr}[0]{\ensuremath{\Hmat^{(r)}}}               
\newcommand{\hvecr}[1]{\ensuremath{\hvec^{(r, #1)}}}           
\newcommand{\hentr}[1]{\ensuremath{h^{(r, #1)}}}               
\newcommand{\hentgam}[1]{\ensuremath{h^{(\hpgamma, #1)}}}      
\newcommand{\hentalpha}[1]{\ensuremath{h^{(\hpalpha, #1)}}}     

\newcommand{\Cmatr}[0]{\ensuremath{\Cmat^{(\rdot)}}}   
\newcommand{\cveck}[0]{\ensuremath{\cvec^{(k)}}}       
\newcommand{\cvecr}[0]{\ensuremath{\cvec^{(\rdot)}}}   
\newcommand{\sveck}[0]{\ensuremath{\svec^{(k)}}}       
\newcommand{\svecr}[0]{\ensuremath{\svec^{(\rdot)}}}   
\newcommand{\centk}[0]{\ensuremath{c^{(k)}}}           
\newcommand{\centr}[0]{\ensuremath{c^{(\rdot)}}}       
\newcommand{\sentk}[0]{\ensuremath{s^{(k)}}}           
\newcommand{\sentr}[0]{\ensuremath{s^{(\rdot)}}}       
\newcommand{\ubstrut}{\vphantom{\left(1 + \sqrt{5}\,\|\rvecdot\| + \tfrac{5}{3}\,\|\rvecdot\|^2\right)}}
\newcommand{\hpbeta}{\ensuremath{\hp{\beta}}}                  

\newcommand\muGP{\ensuremath{\bar{\mu}}}       
\newcommand\sigmaGP{\ensuremath{\bar{\sigma}}} 

\newcommand{\fvectil}[0]{\noisy{\fvec}}

\newcommand{\rvecdot}[0]{\ensuremath{\dot{\rvec}}}  
\newcommand{\Xmatdot}[0]{\ensuremath{\dot{\Xmat}}}  
\newcommand{\Gmatdot}[0]{\ensuremath{\dot{\Gmat}}}  
\newcommand{\xvecdot}[0]{\ensuremath{\dot{\xvec}}}  

\newcommand\noisy[1]{\ensuremath{\widetilde{#1}}}  

\newcommand{\nx}{\ensuremath{n_x}}           
\newcommand{\ndim}{\ensuremath{n_d}}         
\newcommand{\calO}{\ensuremath{\mathcal{O}}} 

\newcommand{\wvecres}{\ensuremath{\bsym{w}_{\mathrm{res}}}}   

\newcommand{\vvecres}{\ensuremath{\bsym{v}_{\mathrm{res}}}}   
\newcommand{\vres}{\ensuremath{v_{\mathrm{res}}}}   

\newcommand{\vveck}{\ensuremath{\bsym{v}_{\kvec}}}                        
\newcommand{\vveckhalf}{\ensuremath{\bsym{v}^{(1/2)}_{\kvec}}}            
\newcommand{\vvecreshalf}{\ensuremath{\bsym{v}^{(1/2)}_{\mathrm{res}}}}   
\newcommand{\vveconehalf}{\ensuremath{\bsym{v}^{(1/2)}_{\one}}}           
\newcommand{\vvecftilhalf}{\ensuremath{\bsym{v}^{(1/2)}_{\fvectil}}}      

\newcommand{\Psimat}{\ensuremath{\mat{\Psi}}}      

\newcommand{\cmark}{\textcolor{green!70!black}{\ding{51}}}  
\newcommand{\xmark}{\textcolor{red}{\ding{55}}}             

\journal{Advances in Engineering Software}

\begin{document}

\begin{frontmatter}

\title{GpFlash: Maximizing Speed and Minimizing Memory for Gaussian Process Training and Prediction}

\author{Andr\'e Marchildon\fnref{fn:amazon}}

\affiliation{organization={Applied Scientist at Amazon Prime Air},
            city={Seattle},
            country={USA}}

\fntext[fn:amazon]{This work was conducted independently of Amazon.com, Inc.
The views expressed are solely those of the author and do not represent Amazon or its affiliates.
No Amazon proprietary data or systems were used.}
\begin{abstract}

This paper presents GpFlash, a Gaussian process library that is focused on fast training and posterior evaluations, minimal memory usage, and scalability to high-dimensional problems.
GpFlash was compared to the popular GP libraries scikit-learn, GPy, GPflow, GPyTorch, and SMT on a MacBook Pro and the supercomputer Trillium.
GpFlash was found to use less memory and to be faster for training and posterior evaluations relative to the other GP libraries for all cases considered.
For example, when using $\nx = 10^4$ training points in a 10-dimensional space GpFlash required 9.6 to 35 times less memory and was 2.6 to 54 times faster than the other GP libraries for the evaluation of the marginal log likelihood and its gradient.
The memory reduction was achieved by only using one matrix of size $\nx \times \nx$ to store different symmetric matrices in its lower and upper triangular entries.
The computational speedup was the result of designing algorithms to access memory efficiently, using BLAS and LAPACK methods extensively, and factoring kernel equations so that expensive exponential calculations could be reused.
GpFlash was more than 440 times faster than all of the other GP libraries considered when evaluating the posterior mean together with its gradient for the case with $10^4$ training points and 10 dimensions.
GpFlash is available at \url{https://github.com/marchildon/gpflash}.

\end{abstract}


\end{frontmatter}

\bibliographystyle{elsarticle-num}
\section{Introduction} \label{Sec_Intro}

Gaussian processes (GPs) are a popular surrogate method that can quantify the uncertainty in their predictions \cite{rasmussen_gaussian_2006}.
A strength of GPs is the wide range of data that they can use, which includes noisy data \cite{ameli_noise_2022,shekhar_signicance_2021} and also gradients \cite{morris_bayesian_1993,dalbey_efficient_2013,laurent_overview_2019,shende_systematic_2022}.
A common application of GPs is Bayesian optimization, usually global optimization \cite{jones_efficient_1998,shahriari_taking_2016}, but also local optimization \cite{eriksson_scalable_2019,marchildon_efficient_2026}, and nonlinearly constrained optimization \cite{gardner_bayesian_2014,marchildon_gradient-enhanced_2024,marchildon_framework_2025}.

It is important to select the hyperparameters of a GP carefully in order to have an accurate posterior surrogate \cite{rasmussen_gaussian_2006}.
The most popular method for hyperparameter tuning is to maximize the marginal log likelihood, which can be done without gradients \cite{toal_kriging_2008,toal_development_2011}, but it is generally done with a gradient-based optimizer since this requires fewer iterations \cite{toal_adjoint_2009,ollar_gradient_2017}.
A significant challenge of using GPs with a large number of training points, $\nx$, is that their memory and training costs scale as $\calO(\nx^2)$ and $\calO(\nx^3)$, respectively \cite{rasmussen_gaussian_2006}.

Approximation methods for the calculation of the marginal log likelihood and its gradient have been developed to address the high training cost of GPs \cite{liu_when_2020,wenger_preconditioning_2022}.
These involve approximating the logarithm of the determinant of a symmetric positive definite matrix \cite{dong_scalable_2017}.
The drawback is that these methods only approximate the marginal log likelihood and its gradient \cite{lin_stochastic_2024}, and most gradient-based optimizers struggle with inexact gradients \cite{nocedal_numerical_2006,martins_engineering_2021}.

GPUs can also accelerate the $\calO(\nx^3)$ computations \cite{wang_exact_2019,gardner_gpytorch_2021}, but they are not as widely available as CPUs.
For example, the GPUs on MacBooks with M chips are limited to single precision, which is generally insufficient since GPs often have matrices that are significantly ill-conditioned \cite{davis_six_1997,kostinski_condition_2000}.

This paper develops algorithms that are faster and use less memory to evaluate the marginal log likelihood and its gradient exactly on CPUs compared to existing GP libraries.
Together they form the open-source library GpFlash: Gaussian Process -- Fast Likelihood, Analytic, Storage-efficient, and High-dimensional.
GpFlash is compared to the GP libraries scikit-learn \cite{pedregosa_scikit-learn_2018}, GPy \cite{gpy_gpy_2012}, GPflow \cite{matthews_gpflow_2016}, GPyTorch \cite{gardner_gpytorch_2021}, and SMT \cite{saves_smt_2024}.
GpFlash was found to be the fastest for both training and posterior evaluations.

Sections~\ref{Sec_Notation} and \ref{Sec_GaussianProcess} present the notation and an overview of GPs.
Sections~\ref{Sec_EfficientKernelEval}, \ref{Sec_EfficientCalcTraining}, and \ref{Sec_EfficientCalcPosterior} show how to efficiently evaluate the kernel function, the marginal log likelihood, and the posterior evaluations, respectively.
Section~\ref{Sec_Benchmarking} details the benchmarking methodology before Sections~\ref{Sec_Training} and \ref{Sec_BenchmarkTraining} cover the training studies and library comparisons, respectively.
Section~\ref{Sec_PosteriorEval} covers the posterior evaluations and Section~\ref{Sec_Conclusions} presents the conclusions.

\section{Notation} \label{Sec_Notation}

In this paper integer quantities are represented with $n$ along with a subscript.
For example, $\nx$ and $\ndim$ represent the number of points and dimensions, respectively.
Lowercase Latin and Greek letters are used for scalars, with the exception of $L$, which denotes the likelihood.
Vectors are denoted in the same way as scalars but are bold, \eg $\rvec$.
Likewise, a vector of ones is denoted with $\one$.
Matrices are denoted with uppercase sans-serif Latin and Greek letters, such as $\Xmat \in \mathbb{R}^{\nx \times \ndim}$, which holds the $\nx$ evaluation points in a $\ndim$-dimensional space.
The $i$-th evaluation point is denoted as $\xvec_i$, which is in the $i$-th row of $\Xmat$.

The operator $\odot$ is the Hadamard product for elementwise multiplication
and the function $\sgn(\cdot)$ returns the sign of its input.
Inner products are denoted by $\langle \cdot, \cdot \rangle$, where a subscript $F$ indicates the Frobenius inner product.

A tilde is used to denote noisy quantities (\eg a function evaluation $\tilde{f}_i$ with Gaussian noise), a hat ($\hp{\cdot}$) is used for hyperparameters, a bar is used for posterior evaluations of the GP (\eg $\muGP_f$), and a dot indicates a dimensionless quantity that has been scaled by the hyperparameter vector of inverse length scales $\hpgammavec$, \eg $\xvecdot = \xvec \odot \hpgammavec$.
Training points are represented with $\xvec$, while $\xvec'$ is a point where the posterior is evaluated.

\section{Gaussian process} \label{Sec_GaussianProcess}

\subsection{Gaussian process regression} \label{Sec_GaussianProcess__regression}

Consider a set of $\nx$ points $\xvec_1, \ldots, \xvec_{\nx}$ in a $\ndim$-dimensional space that is organized in a matrix $\Xmat \in \mathbb{R}^{\nx \times \ndim}$.
A function of interest is evaluated at these $\nx$ points resulting in the vector $\fvectil$ that may contain some independent noise with variance $\sigma_f^2$, with components given by
\begin{align}
 \tilde{f}_i &= f(\xvec_i) + \epsilon_i \\
 \epsilon_i &\sim \mathcal{N}(0, \sigma_f^2).\label{Eq_error_f}
\end{align}

A GP requires a mean and kernel function to form a surrogate with the evaluated data.
A constant $\hpbeta$ is typically used for the mean function and its selection is covered in the next subsection.
The kernel function must be symmetric, \ie $k(\xvec, \yvec) = k(\yvec, \xvec)$, and positive semi-definite \cite{rasmussen_gaussian_2006}.
The kernel matrix $\Kmat$, which is formed by evaluating the kernel for all combinations of points stored in the matrix $\Xmat$, is therefore symmetric positive semi-definite:
\begin{equation} \label{Eq_Kmat}
	\Kmat(\Xmat; \hpgammavec) =
	\begin{bmatrix}
		k(\xvec_{1}, \xvec_{1}; \hpgammavec) 	& k(\xvec_{1}, \xvec_{2}; \hpgammavec) 		& \ldots 	& k(\xvec_{1}, \xvec_{\nx}; \hpgammavec) \\
		k(\xvec_{2}, \xvec_{1}; \hpgammavec) 	& k(\xvec_{2}, \xvec_{2}; \hpgammavec) 		& \ldots 	& k(\xvec_{2}, \xvec_{\nx}; \hpgammavec) \\
		\vdots 									& \vdots									& \ddots 	& \vdots \\
		k(\xvec_{\nx}, \xvec_{1}; \hpgammavec) 	& k(\xvec_{\nx}, \xvec_{2}; \hpgammavec) 	& \ldots 	& k(\xvec_{\nx}, \xvec_{\nx}; \hpgammavec)
	\end{bmatrix} \in \mathbb{R}^{\nx \times \nx},
\end{equation}
where $\hpgammavec \in \mathbb{R}_+^{\ndim}$ is a vector of hyperparameters.
A regularized kernel matrix is formed by adding a nugget, $\eta>0$, to the diagonal of $\Kmat$, which makes it positive definite and thus invertible \cite{micchelli_interpolation_1986}.
The covariance matrix is given by
\begin{align} \label{Eq_Sigma}
 \Sigma = \hpsigK^2 \left( \Kmat + \eta \Wmat \right) + \hpstdfval^2 \Idty,
\end{align}
where $\hpsigK^2$ is the hyperparameter for the signal variance, $\Wmat$ is a diagonal matrix with non-negative entries that is typically set to be the identity, and $\hpstdfval^2$ is the hyperparameter for the noise variance that approximates the generally unknown variance $\sigma_f^2$ from \eqref{Eq_error_f}.
The selection of $\eta$ and $\Wmat$ to bound the matrix condition number is covered in Section~\ref{Sec_EfficientCalcTraining__WellConditionedSigma}.

The mean and variance of the posterior are derived from conditional probability and are evaluated with \cite{rasmussen_gaussian_2006}
\begin{align}
	\muGP_f(\xvec')
		&= \hpbeta + \hpsigK^2 \, \kvec(\xvec', \Xmat) \Sigma^{-1} \left(\fvectil - \hpbeta \one \right) \label{Eq_muGp} \\
	\sigmaGP_f^2(\xvec')
		&= \hpsigK^2 \left[ k(\xvec', \xvec') - \hpsigK^{2} \, \kvec(\xvec', \Xmat) \Sigma^{-1} \kvec(\Xmat, \xvec') \right], \label{Eq_sigmaGp}
\end{align}
where $\kvec(\xvec', \Xmat) = \kvec^\top( \Xmat, \xvec')$ and
\begin{equation} \label{Eq_kvec}
	\kvec(\Xmat, \xvec') =
	\begin{bmatrix}
		k(\xvec_{1}, \xvec') \\
		\vdots \\
		k(\xvec_{\nx}, \xvec')
	\end{bmatrix} \in \mathbb{R}^{\nx}.
\end{equation}
The gradients of the mean and variance of the posterior are also commonly used for applications such as Bayesian optimization and are given by
\begin{align}
	\p{\muGP_f}{\xvec'}
		&= \hpsigK^{2} \left(\fvectil - \hpbeta \one \right)^\top \Sigma^{-1} \p{\kvec}{\xvec'} \label{Eq_Post_mean_grad} \\
	\p{\sigmaGP_f^2}{\xvec'}
		&= \hpsigK^{2} \left.\p{k}{\xvec'} \right|_{(\xvec', \xvec')} - 2 \hpsigK^4\,\kvec(\xvec', \Xmat) \Sigma^{-1} \p{\kvec}{\xvec'}, \label{Eq_Post_var_grad}
\end{align}
which are both row vectors.
The Jacobian $\pline{\kvec}{\xvec'}$ is formed by taking elementwise derivatives of the kernel function:
\begin{equation} \label{Eq_kvec_grad_xvec}
	\left.\p{\kvec}{\xvec'} \right|_{(\Xmat, \xvec'; \hpgammavec)}
		=
		\begin{bmatrix}
			\p{\kvec}{x_1'}
			& \p{\kvec}{x_2'}
			& \ldots
			& \p{\kvec}{x_{\ndim}'}
		\end{bmatrix}_{(\Xmat, \xvec'; \hpgammavec)}
		\in \mathbb{R}^{\nx \times \ndim} \quad \calO(\nx \ndim).
\end{equation}

\subsection{Marginal log likelihood} \label{Sec_GaussianProcess__mll}

The common method to select the hyperparameters is to maximize the likelihood function
\begin{equation} \label{Eq_lkd}
	L(\hpgammavec, \hpbeta, \hpsigK^2, \hpstdfval^2; \Xmat, \fvectil)
	= \frac{e^{-\frac{ \left(\fvectil- \hpbeta \one \right)^\top \Sigma^{-1} \left( \fvectil - \hpbeta \one \right) }{2}}}{\sqrt{ (2 \pi)^{\nx} \det \left(\Sigma \right) }}.
\end{equation}
Taking the natural logarithm of the likelihood function simplifies the calculations:
\begin{equation} \label{Eq_ln_lkd_noisy}
	\ln(L) = - \frac{1}{2} \ln \left( \det \left( \Sigma \right) \right) - \frac{1}{2} \left( \fvectil - \hpbeta \one \right)^\top \Sigma^{-1} \left( \fvectil - \hpbeta \one \right) -\frac{\nx}{2} \ln \left( 2\pi \right),
\end{equation}
where the last term is a constant and is thus generally omitted.
By taking the derivative of $\ln(L)$ with respect to $\hpbeta$, setting it to zero, and isolating for $\hpbeta$, we find this closed-form solution that maximizes $\ln(L)$:
\begin{equation} \label{Eq_hpbeta}
	\hpbeta = \frac{ \one^\top \Sigma^{-1} \fvectil}{ \one^\top \Sigma^{-1} \one}.
\end{equation}

A gradient-based optimizer is commonly used to efficiently maximize the marginal log likelihood \cite{ollar_gradient_2017}.
We now consider the derivative of the marginal log likelihood with respect to an arbitrary hyperparameter $\hpalpha$:
\begin{align}
	\p{\ln(L)}{\hpalpha}
		&= -\frac{1}{2} \p{\ln \left( \det \left( \Sigma \right) \right)}{\hpalpha}
			- \frac{1}{2} \wvecres^\top \p{\Sigma^{-1}}{\hpalpha} \wvecres
			+ \wvecres^\top \Sigma^{-1} \left( \one \p{\hpbeta}{\hpalpha} \right) \nonumber \\
		&= -\frac{1}{2} \text{tr} \left( \Sigma^{-1} \p{\Sigma}{\hpalpha} \right)
			+ \frac{1}{2} \vvecres^\top \p{\Sigma}{\hpalpha} \vvecres
			+ \left( \vvecres^\top \one \right) \p{\hpbeta}{\hpalpha} \label{Eq_grad_ln_lkd_SMT} \\
		&= -\frac{1}{2} \text{tr} \left( \Sigma^{-1} \p{\Sigma}{\hpalpha} \right)
			+ \frac{1}{2} \vvecres^\top \p{\Sigma}{\hpalpha} \vvecres, \label{Eq_grad_ln_lkd_noisy}
\end{align}
where $\wvecres = \fvectil - \hpbeta \one$, $\vvecres = \Sigma^{-1} \wvecres$, Jacobi's formula was used for the derivative of $\det \left( \Sigma \right)$, and the property $\pline{\Sigma^{-1}}{\hpalpha} = - \Sigma^{-1} \pline{\Sigma}{\hpalpha} \Sigma^{-1}$ was used.
Some GP libraries, such as SMT, use \eqref{Eq_grad_ln_lkd_SMT} to calculate the derivatives of $\ln(L)$, which contains the derivative $\pline{\hpbeta}{\hpalpha}$.
However, $\vvecres^\top \one$ is zero, as shown in \ref{SecAppendix_dbetaVanish}, so $\pline{\hpbeta}{\hpalpha}$ does not need to be calculated.

\section{Efficient evaluation of the kernel and of its derivatives} \label{Sec_EfficientKernelEval}

\subsection{Stationary kernels} \label{Sec_EfficientKernelEval__StationaryKernels}

Efficient kernel evaluation uses the dimensionless matrix $\Xmatdot$, whose rows are given by
\begin{equation} \label{Eq_Xmatdot_point}
	\xvecdot_i = \xvec_i \odot \hpgammavec \quad \forall \, i \in \{1, \ldots, \nx \}.
\end{equation}
Stationary kernels depend only on the relative distance between two points rather than on their absolute values:
\begin{equation} \label{Eq_rvec}
	\rvec_{ij} = \xvec_i - \xvec_j,
\end{equation}
where posterior evaluations at $\xvec'$ use $\rvec_i = \xvec' - \xvec_i$, so that derivatives with respect to $\xvec'$ and $\rvec_i$ are the same.
The following are popular stationary kernels considered in this paper:
\begin{align}
	\kSE(\rvecdot)
		&= e^{-\frac{1}{2} \| \rvecdot \|^2}
		\label{Eq_kern_SE} \\
	\kMthreehalf(\rvecdot)
		&= \left(1 + \sqrt{3}\,\|\rvecdot\|\right)\,e^{-\sqrt{3}\,\|\rvecdot\|}
		\label{Eq_kern_Mat3f2} \\
	\kM(\rvecdot)
		&= \left(1 + \sqrt{5}\,\|\rvecdot\| + \tfrac{5}{3}\,\|\rvecdot\|^2\right)\,e^{-\sqrt{5}\,\|\rvecdot\|}
		\label{Eq_kern_Mat5f2} \\
	\kRQ(\rvecdot; \hprho)
		&= \left(1 + \frac{\|\rvecdot\|^2}{2\hprho}\right)^{\!-\hprho}
		\label{Eq_kern_RQ} \\
	\kPer(\rvecdot; \hpavec)
		&= \exp\!\left(-2\sum_{d=1}^{\ndim}\hpa_d^2\,\sin^2\!\left(\pi\,\rdot_d\right)\right),
		\label{Eq_kern_Per}
\end{align}
where $\kSE$ is the squared exponential kernel (also called the Gaussian kernel), $\kMthreehalf$ and $\kM$ are Mat\'ern kernels, $\kRQ$ is the rational quadratic kernel with $\hprho > 0$, and $\kPer$ is the periodic kernel with the hyperparameters $\hpavec$.

\subsection{Common terms} \label{Sec_EfficientKernelEval__CommonTerms}

The kernel function is required for the marginal log likelihood, the posterior, and both of their gradients.
These evaluations contain several common terms that can be calculated efficiently with a factorization containing three types of terms: $c$, $h$, and $s$.
Common terms that are used for the kernel evaluation and its derivatives are denoted by $\centk$ and $\centr$, respectively.
Derivative terms with respect to the $\ell$-th entry of the vector $\rvec$ and a hyperparameter vector $\hp{\alphavec}$ are denoted by $\hentr{\ell}$ and $\hentalpha{\ell}$, respectively, where $\hentgam{\ell}$ is used for the case $\hp{\alpha} = \gamma_\ell$.
Finally, $\sentk$ and $\sentr$ are the static terms, which depend only on $\rvecdot$ and are therefore evaluated once per pair of points for the kernel function and for its derivative evaluations, respectively.
Kernels from Section~\ref{Sec_EfficientKernelEval__StationaryKernels} along with their derivatives can be evaluated using this factorized form:
\begin{align}
	k(\xvec, \yvec; \hpgammavec)
		&= \sentk\,\centk \label{Eq_kern_efficient} \\
	\p{k}{r_{\ell}}
		&= \hentr{\ell}\,\centr \label{Eq_kern_der_r_ell_efficient} \\
	\p{k}{\hpgamma_{\ell}}
		&= \hentgam{\ell}\,\centr, \label{Eq_kern_der_gamma_ell_efficient}
\end{align}
where
\begin{equation} \label{Eq_c_rdot}
	\centr = \sentr\,\centk.
\end{equation}
Table~\ref{Tab_HmatCmat} provides the equations for $\sentk$, $\centk$, $\sentr$, $\hentr{\ell}$, and $\hentalpha{\ell}$.
The evaluation of the Mat\'ern $\frac{5}{2}$ kernel and its derivatives is shown as an example:
\begin{align}
	\kM(\rvecdot)
		&= \underbrace{\left(1 + \sqrt{5}\,\|\rvecdot\| + \tfrac{5}{3}\,\|\rvecdot\|^2\right)\ubstrut}_{\sentk}\;
		\underbrace{e^{-\sqrt{5}\,\|\rvecdot\|}\ubstrut}_{\centk}
		\label{Eq_kern_Mat5f2_factored} \\[1.0ex]
	\p{\kM}{\hpgamma_{\ell}}
		&= \underbrace{\hpgamma_\ell\,r_\ell^2\ubstrut}_{\hentgam{\ell}}\;
		\underbrace{-\tfrac{5}{3}\left(1 + \sqrt{5}\,\|\rvecdot\|\right)\ubstrut}_{\sentr}\;
		\underbrace{e^{-\sqrt{5}\,\|\rvecdot\|}\ubstrut}_{\centk}
		\quad \forall \, \ell \in \{1, \ldots, \ndim \} \label{Eq_kern_Mat5f2_der_gamma_ell} \\[1.0ex]
	\p{\kM}{r_{\ell}}
		&= \underbrace{\hpgamma_\ell^2\,r_\ell\ubstrut}_{\hentr{\ell}}\;
		\underbrace{-\tfrac{5}{3}\left(1 + \sqrt{5}\,\|\rvecdot\|\right)\ubstrut}_{\sentr}\;
		\underbrace{e^{-\sqrt{5}\,\|\rvecdot\|}\ubstrut}_{\centk}
		\quad \forall \, \ell \in \{1, \ldots, \ndim \}.\label{Eq_kern_Mat5f2_der_r_ell}
\end{align}
By using the symmetry of the kernel matrix and the factorization presented in this section, the number of exponential evaluations to evaluate the marginal log likelihood and its gradient is reduced from $\nx^2 (\ndim + 1)$ to only $\nx^2/2$.
Likewise, for posterior mean and gradient evaluations the number of exponential evaluations is reduced from $\nx (\ndim + 1)$ to only $\nx$.

\begin{table}[t]
\centering
\scriptsize
\renewcommand{\arraystretch}{1.3}
\setlength{\tabcolsep}{2pt}
\caption{This table provides the required equations to calculate popular stationary kernels efficiently with \eqref{Eq_kern_efficient} and their gradients with respect to $r_{\ell}$ and $\hpgamma_{\ell}$ with \eqref{Eq_kern_der_r_ell_efficient} and \eqref{Eq_kern_der_gamma_ell_efficient}, respectively, where $\theta_{ij\ell} := \pi \hpgamma_\ell r_{ij\ell}$ and $r_{ij\ell} = x_{i \ell} - x_{j \ell}$.
}
\label{Tab_HmatCmat}
\adjustbox{max width=\linewidth}{%
\begin{tabular}{@{} l c c c c c c @{}}
 \toprule
 & & \multicolumn{2}{c}{Kernel evaluation} & \multicolumn{3}{c}{Gradient evaluation} \\
 \cmidrule(lr){3-4} \cmidrule(lr){5-7}
 Kernel & Eq.\ & $\sentk_{ij}$ & $\centk_{ij}$ & $\sentr_{ij}$ & $\hentr{\ell}_{ij}$ & $\hentalpha{\ell}_{ij}$ \\
 \midrule
 Squared exp.\ ($\hpgamma_\ell$)
 & \eqref{Eq_kern_SE}
 & $1$
 & $e^{-\frac{1}{2}\|\rvecdot_{ij}\|^2}$
 & $-1$
 & $\hpgamma_\ell^2\,r_{ij\ell}$
 & $\hpgamma_\ell\,r_{ij\ell}^2$ \\
 \greyrule
 Mat\'ern $\tfrac{5}{2}$ ($\hpgamma_\ell$)
 & \eqref{Eq_kern_Mat5f2}
 & $1+\sqrt{5}\|\rvecdot_{ij}\|+\tfrac{5}{3}\|\rvecdot_{ij}\|^2$
 & $e^{-\sqrt{5}\|\rvecdot_{ij}\|}$
 & $-\tfrac{5}{3}(1+\sqrt{5}\|\rvecdot_{ij}\|)$
 & $\hpgamma_\ell^2\,r_{ij\ell}$
 & $\hpgamma_\ell\,r_{ij\ell}^2$ \\
 \greyrule
 Mat\'ern $\tfrac{3}{2}$ ($\hpgamma_\ell$)
 & \eqref{Eq_kern_Mat3f2}
 & $1+\sqrt{3}\|\rvecdot_{ij}\|$
 & $e^{-\sqrt{3}\|\rvecdot_{ij}\|}$
 & $-3$
 & $\hpgamma_\ell^2\,r_{ij\ell}$
 & $\hpgamma_\ell\,r_{ij\ell}^2$ \\
 \greyrule
 Rational quad.\ ($\hpgamma_\ell$)
 & \multirow{2}{*}{\eqref{Eq_kern_RQ}}
 & \multirow{2}{*}{$1+\tfrac{\|\rvecdot_{ij}\|^2}{2\hprho}$}
 & \multirow{2}{*}{$\left(1+\tfrac{\|\rvecdot_{ij}\|^2}{2\hprho}\right)^{\!-\hprho-1}$}
 & \multirow{2}{*}{$-1$}
 & \multirow{2}{*}{$\hpgamma_\ell^2\,r_{ij\ell}$}
 & $\hpgamma_\ell\,r_{ij\ell}^2$ \\
 Rational quad.\ ($\hprho$)
 &
 &
 &
 &
 &
 & $\sentk_{ij}\ln \sentk_{ij}-\left(\sentk_{ij}-1\right)$ \\
 \greyrule
 Periodic ($\hpgamma_\ell$)
 & \multirow{2}{*}{\eqref{Eq_kern_Per}}
 & \multirow{2}{*}{$1$}
 & \multirow{2}{*}{$\exp\!\left(-2\displaystyle\sum_{d=1}^{\ndim}\hpa_d^2\sin^2(\theta_{ijd})\right)$}
 & \multirow{2}{*}{$1$}
 & \multirow{2}{*}{$-2\pi\,\hpgamma_\ell\,\hpa_\ell^2\sin(2 \theta_{ij\ell})$}
 & $-2\pi\,r_{ij\ell}\,\hpa_\ell^2\sin(2 \theta_{ij\ell})$ \\
 Periodic ($\hpa_\ell$)
 &
 &
 &
 &
 &
 & $-4\hpa_\ell \sin^2 \theta_{ij\ell}$ \\
 \bottomrule
\end{tabular}%
}
\end{table}

\subsection{BLAS with the polarization identity} \label{Sec_EfficientKernelEval__BLAS}

The polarization identity can be used for the following common calculation for stationary kernels:
\begin{equation} \label{Eq_diff_xvec_property}
	\| \rvecdot_{ij} \|^2 = \|\xvecdot_i\|^2 + \|\xvecdot_j\|^2 - 2 \xvecdot_i^\top \xvecdot_j.
\end{equation}
This relation has to be applied to each pair of points in $\Xmatdot$ when forming the kernel matrix $\Kmat$ from \eqref{Eq_Kmat}.
This can be done efficiently by calculating the norm of the $\nx$ dimensionless points once and forming the Gram matrix
\begin{equation} \label{Eq_Gmat}
	\Gmatdot = \Xmatdot \Xmatdot^\top \quad \calO(\nx^2 \ndim),
\end{equation}
which can be calculated with the BLAS method \texttt{DGEMM}.
GpFlash makes use of \texttt{DSYRK} since it only requires the lower half entries of $\Gmatdot$.

\section{Efficient training calculations} \label{Sec_EfficientCalcTraining}

\subsection{Forming a well-conditioned covariance matrix} \label{Sec_EfficientCalcTraining__WellConditionedSigma}

The covariance matrix, $\Sigma$, can be severely ill-conditioned and this gets worse as the training points get closer together \cite{marchildon_non-intrusive_2023}.
Having an ill-conditioned matrix can cause the Cholesky decomposition to fail, and it increases the iterations required for iterative solvers to achieve a desired tolerance.

The ill-conditioning of $\Sigma$ from \eqref{Eq_Sigma} can be addressed by using $\eta > 0$ \cite{rasmussen_gaussian_2006}.
However, there is a balance between having $\eta$ large enough to address the ill-conditioning problem, but not so large that it degrades the accuracy of the posterior.
In this subsection a simplified method from \cite{marchildon_solution_2024} is presented to address the ill-conditioning of a covariance matrix formed using a stationary kernel.
For stationary kernels, $\Kmat$ is a correlation matrix and we can ensure that $\kappa(\Sigma) \leq \condmax$ with $\Wmat=\Idty$ from \eqref{Eq_Sigma} and
\begin{alignat}{2}
	\eta
		&= \frac{1 + \max_i \sum_{\substack{j=1 \\ j \neq i}}^{\nx} \left| k_{ij} \right|}{\condmax -1} \quad
		&&\calO(\nx^2) \label{Eq_eta} \\
		&\leq \frac{\nx}{\condmax -1}
		&&\calO(1), \label{Eq_eta_upper_bound}
\end{alignat}
where the numerator of \eqref{Eq_eta} is an upper bound on the maximum eigenvalue of $\Kmat$ derived from the Gershgorin circle theorem.
The upper bound for $\eta$ from \eqref{Eq_eta_upper_bound} is derived by assuming each entry in $\Kmat$ is equal to one.
For non-stationary kernels, $\Kmat$ in \eqref{Eq_eta} is replaced by a preconditioned matrix and $\Wmat$ from \eqref{Eq_Sigma} might not be equal to the identity matrix \cite{marchildon_solution_2024}.

\begin{table}[t]
\centering
\renewcommand{\arraystretch}{1.3}
\caption{All of the GP libraries considered in this table use the Cholesky decomposition.
GPyTorch, however, uses a conjugate gradient solver when $\nx > 800$, and adds a pivoted Cholesky preconditioner once $\nx \geq 2000$.
The Cholesky decomposition for GpFlash is done on a matrix with a bounded condition number since it uses the method from \cite{marchildon_solution_2024}.
}
\label{Tab_linear_solve}
\adjustbox{max width=\linewidth}{%
\begin{tabular}{lcccccc}
 \toprule
 & scikit-learn & GPy & GPyTorch & GPflow & SMT & GpFlash \\
 \midrule
 Default $\eta$ & $10^{-10}$ &$0$ & $0$ & $10^{-6}$ & $2.2 \times 10^{-14}$ & Eq.\ \eqref{Eq_eta} \\
 Retry & \xmark & 5 retries: $10^{k-6} \, \frac{\text{tr}(\Sigma)}{\nx}$ & 3 retries: $10^{k-8}$ & \xmark & \xmark & Not needed \\
	Ensures $\kappa(\Sigma) \leq \condmax$ & \xmark & \xmark & \xmark & \xmark & \xmark & \cmark \\
 \bottomrule
\end{tabular}%
}
\end{table}

Table~\ref{Tab_linear_solve} indicates how the other GP libraries considered in this paper address the ill-conditioning problem of the covariance matrix $\Sigma$.
The libraries scikit-learn, GPflow, and SMT set a default $\eta$ value that is independent of $\nx$ and of the entries in $\Kmat$.
This often results in a nugget value larger than needed for $\Sigma$ to be well conditioned, while in other cases it will be insufficient and the Cholesky factorization will fail.
GPy and GPyTorch add no nugget on the first attempt and, on a Cholesky failure, retry with an incrementally larger nugget.
This is computationally expensive and still does not ensure that the final attempt succeeds.

\subsection{Efficient implementation of the log likelihood calculations} \label{Sec_EfficientCalcTraining__Lkd}

The matrix $\Sigma$ from \eqref{Eq_Sigma} must be reformed each time the hyperparameters are updated during the optimization process.
This cost scales as $\calO(\nx^2 \ndim)$, where the dependence on $\nx$ is generally stronger than on $\ndim$ since the number of exponential evaluations only depends on the former.
The Cholesky factor, $\Lmat$, is required for several later calculations:
\begin{equation} \label{Eq_Lmat}
	\Lmat \Lmat^\top = \Sigma \quad \calO(\nx^3).
\end{equation}
The closed-form solution for $\hpbeta$ from \eqref{Eq_hpbeta} is evaluated at a cost of $\calO(\nx^2)$ using two triangular half solves with
\begin{alignat}{2}
	\vveconehalf
		&= \Lmat^{-1} \one \quad
		&& \calO(\nx^2) \label{Eq_vveconehalf}\\
	\vvecftilhalf
		&= \Lmat^{-1} \fvectil \quad
		&& \calO(\nx^2) \label{Eq_vvecftilhalf}\\
	\hpbeta
		&= \frac{\langle \vvecftilhalf, \, \vveconehalf \rangle}{\| \vveconehalf \|^2} \quad
		&& \calO(\nx).\label{Eq_hpbeta_efficient}
\end{alignat}
The marginal log likelihood, $\ln(L)$, from \eqref{Eq_ln_lkd_noisy} is evaluated with:
\begin{alignat}{2}
	 \ln \left( \det \left( \Sigma \right) \right)
		&= 2 \sum_{i=1}^{\nx} \ln \left( \Lmat_{ii} \right) \quad
		&& \calO(\nx) \label{Eq_ln_det_Sigma}\\
	\vvecreshalf
		&= \vvecftilhalf - \hpbeta\, \vveconehalf \quad
		&& \calO(\nx) \label{Eq_half_res_solve} \\
	\ln(L)
		&= -\frac{1}{2} \ln \left( \det \left( \Sigma \right) \right) - \frac{1}{2} \| \vvecreshalf \|^2 \quad
		&& \calO(\nx), \label{Eq_ln_lkd_noisy_efficient}
\end{alignat}
where the constant term from \eqref{Eq_ln_lkd_noisy} has been omitted since it does not impact the selection of the hyperparameters.

\subsection{Calculating the gradient of $\ln(L)$ with the Frobenius inner product} \label{Sec_EfficientCalcTraining__Grad_lkd}

For gradient calculations the first vector that must be calculated is
\begin{equation} \label{Eq_vvecres}
	\vvecres = \Lmat^{-\top} \vvecreshalf \quad \calO(\nx^2),
\end{equation}
where $\vvecreshalf$ comes from \eqref{Eq_half_res_solve}.
Using \eqref{Eq_grad_ln_lkd_noisy} to calculate the derivative of the marginal log likelihood is inefficient since it involves $\Sigma^{-1} \pline{\Sigma}{\hpalpha}$, which scales as $\calO(\nx^3)$, so the full gradient with respect to $\hpgammavec$ would cost $\calO(\nx^3 \ndim)$.
The expensive matrix-matrix products are eliminated with \cite{rasmussen_gaussian_2006}:
\begin{align}
	\p{\ln(L)}{\hpalpha}
		&= 	-\frac{1}{2} \text{tr} \left( \Sigma^{-1} \p{\Sigma}{\hpalpha} \right)
			+\frac{1}{2} \text{tr} \left(\vvecres^\top \p{\Sigma}{\hpalpha} \vvecres \right) \nonumber \\
		&= 	-\frac{1}{2} \text{tr} \left( \Sigma^{-1} \p{\Sigma}{\hpalpha} \right)
			+\frac{1}{2} \text{tr} \left(\p{\Sigma}{\hpalpha} \vvecres \vvecres^\top \right) \nonumber \\
		&= \langle \Psimat,\, \p{\Sigma}{\hpalpha} \rangle_F \quad \calO(\nx^2), \label{Eq_der_lkd_efficient}
\end{align}
where $\vvecres$ is from \eqref{Eq_vvecres}, the cyclic and linear properties of the trace operator were used, $\langle \cdot, \cdot \rangle_F$ is the Frobenius inner product, and
\begin{equation} \label{Eq_Psimat}
	\Psimat = \frac{1}{2} \left( \vvecres\vvecres^\top - \Sigma^{-1} \right) \quad \calO(\nx^3).
\end{equation}
It is generally not recommended to form the explicit inverse of a matrix such as $\Sigma^{-1}$ since it is expensive and can lead to large numerical errors when the matrix is ill-conditioned \cite{chapra_numerical_2015}.
However, the computational cost of forming $\Psimat$ is offset by the lower cost of using \eqref{Eq_der_lkd_efficient} compared to \eqref{Eq_grad_ln_lkd_noisy}, while the ill-conditioning problem is addressed by using the method presented in Section~\ref{Sec_EfficientCalcTraining__WellConditionedSigma}.
The use of \eqref{Eq_der_lkd_efficient} with $\hpalpha = \hpgamma_\ell$ for $\ell \in \{1, \ldots, \ndim \}$ requires
\begin{alignat}{2}
	\p{\eta}{\hpgamma_\ell}
		&= \frac{1}{\condmax -1} \sum_{\substack{j=1 \\ j \neq i^\star}}^{\nx} \sgn \left( k_{i^\star j} \right) \p{k_{i^\star j}}{\hpgamma_\ell} \quad
		&& \calO(\nx) \label{Eq_der_eta_wrt_gamma} \\
	\p{\Sigma}{\hpgamma_\ell}
		&= \hpsigK^2 \left( \p{\Kmat}{\hpgamma_\ell} + \p{\eta}{\hpgamma_\ell} \Wmat + \eta \p{\Wmat}{\hpgamma_\ell} \right) \nonumber \\
		&= \hpsigK^2 \left( \p{\Kmat}{\hpgamma_\ell} + \p{\eta}{\hpgamma_\ell} \Wmat \right) \quad
		&&\calO(\nx^2), \label{Eq_der_Sigma_wrt_gamma}
\end{alignat}
where $\Wmat$ is constant for stationary kernels and $i^\star$ is the index of the row of $\Kmat$ that provides the maximum for $\eta$ from \eqref{Eq_eta}.
The derivatives of $\ln(L)$ with respect to $\hprho$ and the entries of $\hpavec$ for the rational quadratic and periodic kernels, respectively, are analogous to \eqref{Eq_der_eta_wrt_gamma} and \eqref{Eq_der_Sigma_wrt_gamma}.
The gradient of the marginal log likelihood with respect to $\hpgammavec$ is calculated with
\begin{equation} \label{Eq_gradient_ln_L_with_frobenius_Psimat}
	\p{\ln(L)}{\hpgammavec} =  \hpsigK^2 \left[
		\langle \Psimat,\, \p{\Kmat}{\hpgamma_1} \rangle_F + \zeta \p{\eta}{\hpgamma_1},
		\ldots,
		\langle \Psimat,\, \p{\Kmat}{\hpgamma_{\ndim}} \rangle_F + \zeta \p{\eta}{\hpgamma_{\ndim}}
		\right] \calO(\nx^2 \ndim),
\end{equation}
where
\begin{equation} \label{Eq_zeta}
	\zeta = \sum_{i=1}^{\nx} \Psimat_{ii} \Wmat_{ii} \quad \calO(\nx).
\end{equation}

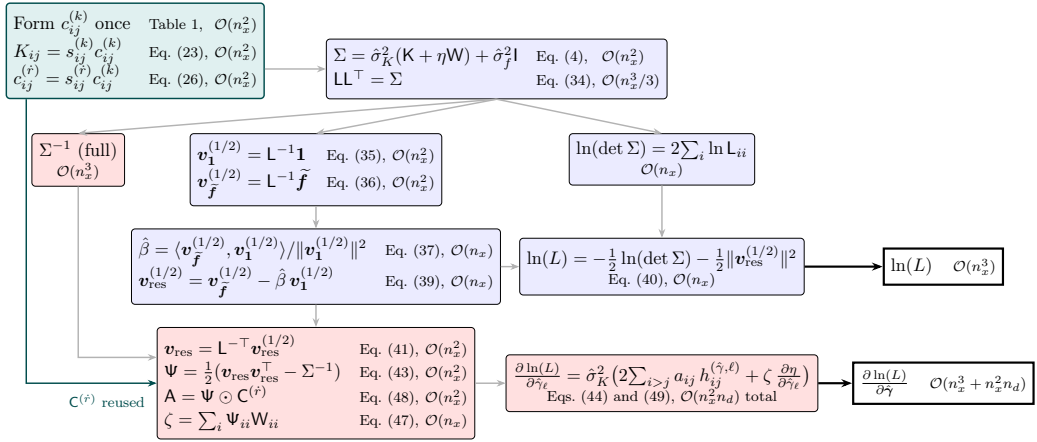
\begin{figure}[t]
\centering
\resizebox{\textwidth}{!}{
\begin{tikzpicture}[
    box/.style={rectangle, draw, rounded corners=2pt,
                minimum height=0.7cm, align=center, font=\footnotesize},
    valbox/.style={box, fill=blue!8},
    gradbox/.style={box, fill=red!12},
    sharebox/.style={box, fill=teal!12, draw=teal!60!black, thick},
    outbox/.style={rectangle, draw=black, very thick,
                    minimum height=0.7cm, align=center,
                    font=\footnotesize\bfseries, fill=white},
    arrow/.style={-{Stealth[length=5pt]}, thick, gray!60},
    outarrow/.style={-{Stealth[length=6pt]}, very thick, black},
    reusearrow/.style={-{Stealth[length=5pt]}, thick, teal!60!black},
]
\node[sharebox, anchor=south, align=left] (formK) at (2,-0.5)
    {\begin{tabular}{@{}l@{\quad}l@{\ }l@{}}
     Form $c^{(k)}_{ij}$ once & {\scriptsize Table~\ref{Tab_HmatCmat},} & {\scriptsize $\calO(\nx^2)$}\\[1pt]
     $K_{ij}=s^{(k)}_{ij}c^{(k)}_{ij}$ & {\scriptsize Eq.~\eqref{Eq_kern_efficient},} & {\scriptsize $\calO(\nx^2)$}\\[1pt]
     $c^{(\rdot)}_{ij}=s^{(\rdot)}_{ij}c^{(k)}_{ij}$ & {\scriptsize Eq.~\eqref{Eq_c_rdot},} & {\scriptsize $\calO(\nx^2)$}
     \end{tabular}};
\node[valbox, anchor=south, align=left] (chol) at (9,-0.5)
    {\begin{tabular}{@{}l@{\quad}l@{\ }l@{}}
     $\Sigma = \hpsigK^2(\Kmat + \eta\Wmat) + \hpstdfval^2\Idty$ & {\scriptsize Eq.~\eqref{Eq_Sigma},} & {\scriptsize $\calO(\nx^2)$}\\[1pt]
     $\Lmat\Lmat^\top = \Sigma$ & {\scriptsize Eq.~\eqref{Eq_Lmat},} & {\scriptsize $\calO(\nx^3 / 3)$}
     \end{tabular}};

\node[valbox, anchor=north, align=left] (halfsolves) at (5.5,-1.16)
    {\begin{tabular}{@{}l@{\quad}l@{\ }l@{}}
     $\vveconehalf = \Lmat^{-1}\one$ & {\scriptsize Eq.~\eqref{Eq_vveconehalf},} & {\scriptsize $\calO(\nx^2)$}\\[1pt]
     $\vvecftilhalf = \Lmat^{-1}\fvectil$ & {\scriptsize Eq.~\eqref{Eq_vvecftilhalf},} & {\scriptsize $\calO(\nx^2)$}
     \end{tabular}};
\node[valbox, anchor=north] (logdet) at (12.2,-1.16)
    {$\ln(\det\Sigma) = 2\!\sum_i\ln\Lmat_{ii}$\\[-1pt]{\scriptsize $\calO(\nx)$}};

\node[gradbox, align=left] (psi) at (5.5,-6)
    {\begin{tabular}{@{}l@{\quad}l@{\ }l@{}}
     $\vvecres = \Lmat^{-\top}\vvecreshalf$ & {\scriptsize Eq.~\eqref{Eq_vvecres},} & {\scriptsize $\calO(\nx^2)$}\\[1pt]
     $\Psimat = \tfrac12(\vvecres\vvecres^\top - \Sigma^{-1})$ & {\scriptsize Eq.~\eqref{Eq_Psimat},} & {\scriptsize $\calO(\nx^2)$}\\[1pt]
     $\Amat = \Psimat \odot \Cmatr$ & {\scriptsize Eq.~\eqref{Eq_Amat},} & {\scriptsize $\calO(\nx^2)$}\\[1pt]
     $\zeta = \sum_i \Psimat_{ii}\Wmat_{ii}$ & {\scriptsize Eq.~\eqref{Eq_zeta},} & {\scriptsize $\calO(\nx)$}
     \end{tabular}};
\node[gradbox] (gradd) at (12.2,-6)
    {$\p{\ln(L)}{\hpgamma_\ell} = \hpsigK^2\bigl(2\!\sum_{i>j} a_{ij}\,\hentgam{\ell}_{ij} + \zeta\,\p{\eta}{\hpgamma_\ell}\bigr)$\\[-1pt]{\scriptsize Eqs.~\eqref{Eq_der_eta_wrt_gamma} and \eqref{Eq_gradient_ln_L_with_frobenius_Amat},\ $\calO(\nx^2 \ndim)$ total}};
\node[outbox] (gradmll) at (17.65,-6)
    {$\p{\ln(L)}{\hpgammavec}$\quad{\scriptsize $\calO(\nx^3 + \nx^2\ndim)$}};

\node[valbox, align=left] (bhat) at ($(halfsolves.south)!0.5!(psi.north)$)
    {\begin{tabular}{@{}l@{\quad}l@{\ }l@{}}
     $\hpbeta = \langle \vvecftilhalf, \vveconehalf \rangle / \| \vveconehalf \|^{2}$ & {\scriptsize Eq.~\eqref{Eq_hpbeta_efficient},} & {\scriptsize $\calO(\nx)$}\\[1pt]
     $\vvecreshalf = \vvecftilhalf - \hpbeta\,\vveconehalf$ & {\scriptsize Eq.~\eqref{Eq_half_res_solve},} & {\scriptsize $\calO(\nx)$}
     \end{tabular}};
\node[valbox] (mll) at (bhat -| logdet)
    {$\ln(L) = -\tfrac12\ln(\det\Sigma) - \tfrac12\|\vvecreshalf\|^2$\\[-1pt]{\scriptsize Eq.~\eqref{Eq_ln_lkd_noisy_efficient},\ $\calO(\nx)$}};
\node[outbox] (mllout) at (bhat -| gradmll)
    {$\ln(L)$\quad{\scriptsize $\calO(\nx^3)$}};

\node[gradbox, anchor=north east] (siginv) at ($(bhat.west |- halfsolves.north)+(-0.12,0)$)
    {$\Sigma^{-1}$ (full)\\[-1pt]{\scriptsize $\calO(\nx^3)$}};

\draw[arrow] (formK.east |- chol.west) -- (chol.west);
\draw[arrow] (chol.south)   -- (siginv.north);
\draw[arrow] (chol.south)   -- (halfsolves.north);
\draw[arrow] (chol.south)   -- (logdet.north);
\draw[arrow] (logdet.south) -- (mll.north);
\draw[arrow] (halfsolves.south) -- (bhat.north);
\draw[arrow] (bhat.east)    -- (mll.west);
\draw[arrow] (bhat.south)   -- (bhat.south |- psi.north);
\draw[arrow] (siginv.south) |- ($(psi.west)+(0,0.5)$);
\draw[arrow] (psi.east)     -- (gradd.west);
\draw[outarrow] (mll.east)   -- (mllout.west);
\draw[outarrow] (gradd.east) -- (gradmll.west);

\draw[reusearrow] ($(formK.south west)+(0.4,0)$) |- (psi.west);
\node[font=\scriptsize, text=teal!60!black, anchor=north] at (1.45,-6.06)
    {$\Cmatr$ reused};
\end{tikzpicture}}
\caption{Flowchart for calculations needed for the marginal log likelihood $\ln(L)$ (blue boxes) and its gradient (red boxes) with respect to $\hpgammavec$, as implemented in GpFlash.}
\label{Fig_MLLPipeline}
\end{figure}

The computational cost of calculating the gradient using the Frobenius inner product can be reduced further by first calculating
\begin{equation} \label{Eq_Amat}
	\Amat = \Psimat \odot \Cmatr \quad \calO(\nx^2),
\end{equation}
where $\Psimat$ comes from \eqref{Eq_Psimat}, $\Cmatr$ from \eqref{Eq_c_rdot}, and the kernel-specific equations are in Table~\ref{Tab_HmatCmat}.
The gradient of the marginal log likelihood from \eqref{Eq_gradient_ln_L_with_frobenius_Psimat} can thus be calculated more efficiently with
\begin{equation} \label{Eq_gradient_ln_L_with_frobenius_Amat}
	\p{\ln(L)}{\hpgammavec} = \hpsigK^2 \left[
		\langle \Amat,\, \Hmatgam{1} \rangle_F + \zeta \p{\eta}{\hpgamma_1},
		\ldots,
		\langle \Amat,\, \Hmatgam{\ndim} \rangle_F + \zeta \p{\eta}{\hpgamma_{\ndim}} \right] \calO(\nx^2 \ndim).
\end{equation}
The use of \eqref{Eq_gradient_ln_L_with_frobenius_Psimat} or \eqref{Eq_gradient_ln_L_with_frobenius_Amat} requires the same number of exponential evaluations if the former uses \eqref{Eq_kern_der_gamma_ell_efficient} to calculate the derivative matrices.
However, \eqref{Eq_gradient_ln_L_with_frobenius_Amat} is more efficient since it requires fewer multiplications, as the entries of $\Cmatr$ are only multiplied with $\Psimat$, to form $\Amat$, rather than with $\ndim$ matrices.
Two additional derivative calculations that are needed are
\begin{align}
	\p{\ln(L)}{\hpstdfval^2}
		&= \text{tr} \left( \Psimat \right) \label{Eq_der_ln_lkd_wrt_sigf2} \\
	\p{\ln(L)}{\hpsigK^2}
		&= \langle \Psimat,\, \Kmat \rangle_F + \eta \zeta, \label{Eq_der_ln_lkd_wrt_sigK2}
\end{align}
where $\zeta$ comes from \eqref{Eq_zeta} and the relations $\pline{\Sigma}{\hpsigK^2} = \Kmat + \eta \Wmat$ and $\pline{\Sigma}{\hpstdfval^2} = \Idty$ were used.
Figure~\ref{Fig_MLLPipeline} shows the calculations required for the marginal log likelihood (blue boxes) and its gradient (red boxes).

\section{Efficient posterior evaluations for stationary kernels} \label{Sec_EfficientCalcPosterior}

\subsection{A priori calculations} \label{Sec_EfficientCalcPosterior__APriori}

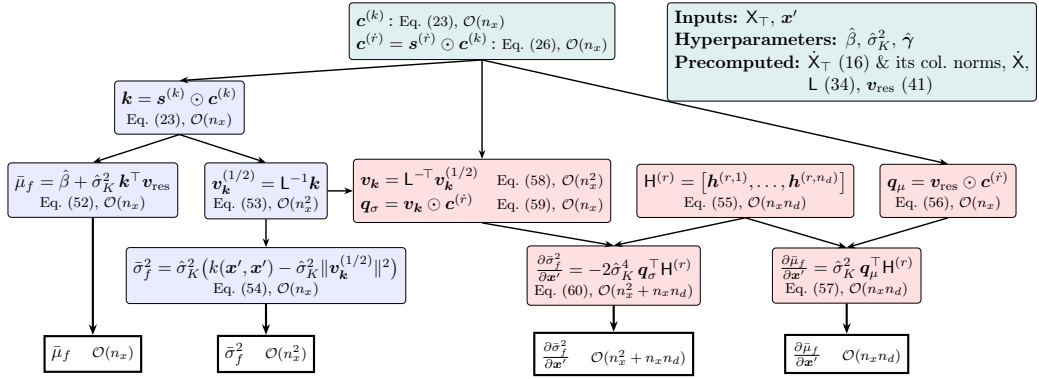
\begin{figure}[t]
\centering
\resizebox{\textwidth}{!}{
\begin{tikzpicture}[
    node distance=0.55cm and 1.3cm,
    box/.style={rectangle, draw, rounded corners=2pt,
                minimum height=0.7cm, align=center, font=\footnotesize},
    valbox/.style={box, fill=blue!8},
    gradbox/.style={box, fill=red!12},
    sharebox/.style={box, fill=teal!12, align=left},
    inputbox/.style={box, fill=teal!12, align=left},
    outbox/.style={rectangle, draw=black, very thick,
                    minimum height=0.7cm, align=center,
                    font=\footnotesize\bfseries, fill=white},
    arrow/.style={-{Stealth[length=5pt]}, thick, black},
    outarrow/.style={-{Stealth[length=6pt]}, very thick, black},
]
\node[valbox] (kstar) at (1.0,0)
    {$\kvec = \sveck \odot \cveck$\\[-1pt]{\scriptsize Eq.~\eqref{Eq_kern_efficient},\ $\calO(\nx)$}};
\node[valbox, below=0.55cm of kstar, xshift=-1.7cm] (mu)
    {$\muGP_f = \hpbeta + \hpsigK^2\,\kvec^\top\vvecres$\\[-1pt]{\scriptsize Eq.~\eqref{Eq_muGp_efficient},\ $\calO(\nx)$}};
\node[valbox, below=0.55cm of kstar, xshift=1.7cm] (V)
    {$\vveckhalf = \Lmat^{-1}\kvec$\\[-1pt]{\scriptsize Eq.~\eqref{Eq_vveckhalf},\ $\calO(\nx^2)$}};
\node[valbox, below=0.55cm of V] (sig2)
    {$\sigmaGP_f^2 = \hpsigK^2\big(k(\xvec',\xvec') - \hpsigK^2\|\vveckhalf\|^2\big)$\\[-1pt]{\scriptsize Eq.~\eqref{Eq_sigmaGp_efficient},\ $\calO(\nx)$}};
\node[gradbox, right=0.5cm of V, align=left] (Vp)
    {\begin{tabular}{@{}l@{\quad}l@{\ }l@{}}
     $\vveck = \Lmat^{-\!\top}\vveckhalf$ & {\scriptsize Eq.~\eqref{Eq_vvec_kvec},} & {\scriptsize $\calO(\nx^2)$}\\[1pt]
     $\qvec_{\sigma} = \vveck \odot \cvecr$ & {\scriptsize Eq.~\eqref{Eq_qvec_sigma},} & {\scriptsize $\calO(\nx)$}
     \end{tabular}};

\node[gradbox, right=0.5cm of Vp] (Hmat)
    {$\Hmatr = \left[\hvecr{1}, \ldots, \hvecr{\ndim}\right]$\\[-1pt]{\scriptsize Eq.~\eqref{Eq_Hmat_r},\ $\calO(\nx\ndim)$}};
\node[gradbox, right=0.5cm of Hmat] (dkstar)
    {$\qvec_{\mu} = \vvecres \odot \cvecr$\\[-1pt]{\scriptsize Eq.~\eqref{Eq_qvec_mu},\ $\calO(\nx)$}};
\coordinate (midsig) at ($(Vp)!0.5!(Hmat)$);
\coordinate (midmu)  at ($(Hmat)!0.5!(dkstar)$);
\node[gradbox] (dsig2) at (midsig |- sig2)
    {$\p{\sigmaGP_f^2}{\xvec'} = -2\hpsigK^4\, \qvec_{\sigma}^\top \Hmatr$\\[-1pt]{\scriptsize Eq.~\eqref{Eq_Post_var_grad_efficient},\ $\calO(\nx^2 + \nx\ndim)$}};
\node[gradbox] (dmu) at (midmu |- sig2)
    {$\p{\muGP_f}{\xvec'} = \hpsigK^2 \, \qvec_{\mu}^\top \Hmatr$\\[-1pt]{\scriptsize Eq.~\eqref{Eq_Post_mean_grad_efficient},\ $\calO(\nx\ndim)$}};

\node[sharebox, yshift=1.5cm] (ck) at (Vp |- kstar)
    {$\cveck$\,:\ {\scriptsize Eq.~\eqref{Eq_kern_efficient},\ $\calO(\nx)$}\\[1pt]
     $\cvecr = \svecr \odot \cveck$\,:\ {\scriptsize Eq.~\eqref{Eq_c_rdot},\ $\calO(\nx)$}};

\node[inputbox, anchor=north east, xshift=4mm] (inputs) at (dkstar.east |- ck.north)
    {\textbf{Inputs:} $\Xmat_\top$, $\xvec'$\\
     \textbf{Hyperparameters:} $\hpbeta$, $\hpsigK^2$, $\hpgammavec$\\
     \textbf{Precomputed:} $\Xmatdot_\top$~\eqref{Eq_Xmatdot_point} \& its col.\ norms, $\Xmatdot$,\\
     \hphantom{\textbf{Precomputed:} }$\Lmat$~\eqref{Eq_Lmat}, $\vvecres$~\eqref{Eq_vvecres}};

\path (dsig2.south) ++(0,-0.95cm) coordinate (out_y);
\node[outbox] (muout)    at (mu    |- out_y) {$\muGP_f$\quad{\scriptsize $\calO(\nx)$}};
\node[outbox] (sig2out)  at (sig2  |- out_y) {$\sigmaGP_f^2$\quad{\scriptsize $\calO(\nx^2)$}};
\node[outbox] (dsig2out) at (dsig2 |- out_y) {$\p{\sigmaGP_f^2}{\xvec'}$\quad{\scriptsize $\calO(\nx^2 + \nx\ndim)$}};
\node[outbox] (dmuout)   at (dmu   |- out_y) {$\p{\muGP_f}{\xvec'}$\quad{\scriptsize $\calO(\nx\ndim)$}};

\draw[arrow] (ck.south) -- (kstar.north);
\draw[arrow] (ck.south) -- (dkstar.north);
\draw[arrow] (ck.south) -- (Vp.north);
\draw[arrow] (kstar.south)  -- (mu.north);
\draw[arrow] (kstar.south)  -- (V.north);
\draw[arrow] (V.south)      -- (sig2.north);
\draw[arrow] (V.east)       -- (Vp.west);
\draw[arrow] (dkstar.south) -- (dmu.north);
\draw[arrow] (Hmat.south)   -- (dmu.north);
\draw[arrow] (Hmat.south)   -- (dsig2.north);
\draw[arrow] (Vp.south)     -- (dsig2.north);
\draw[outarrow] (mu.south)    -- (muout.north);
\draw[outarrow] (sig2.south)  -- (sig2out.north);
\draw[outarrow] (dsig2.south) -- (dsig2out.north);
\draw[outarrow] (dmu.south)   -- (dmuout.north);
\end{tikzpicture}}
\caption{This flowchart shows how the posterior mean and variance (blue boxes) and their gradients (red boxes) are calculated efficiently.}
\label{Fig_PosteriorPipelineEfficient}
\end{figure}

Figure~\ref{Fig_PosteriorPipelineEfficient} shows how the posterior can be evaluated efficiently using the process presented in this section.
The Cholesky factor $\Lmat$ for the matrix $\Sigma$ is calculated with \eqref{Eq_Lmat} along with $\vvecres$ from \eqref{Eq_vvecres}.
The dimensionless matrix $\Xmatdot$ from \eqref{Eq_Xmatdot_point} is calculated a priori along with the squared $\ell_2$ norms of its rows so that the polarization identity from Section~\ref{Sec_EfficientKernelEval__BLAS} can be used.

\subsection{Posterior mean and variance evaluations} \label{Sec_EfficientCalcPosterior__MeanVar}

The mean of the posterior is evaluated at $\xvec'$ using
\begin{equation} \label{Eq_muGp_efficient}
	\muGP_f(\xvec')
		= \hpbeta + \hpsigK^2 \, \kvec^\top \vvecres
		\quad \calO(\nx),
\end{equation}
where $\kvec$ is formed with \eqref{Eq_kern_efficient} for a stationary kernel and \eqref{Eq_kvec} otherwise.
The variance of the posterior is calculated with
\begin{alignat}{2}
	\vveckhalf
		&= \Lmat^{-1} \kvec \quad
		&&\calO(\nx^2) \label{Eq_vveckhalf} \\
	\sigmaGP_f^2(\xvec')
		&= \hpsigK^2 \left( k(\xvec', \xvec') - \hpsigK^{2} \, \| \vveckhalf \|^2 \right) \quad
		&& \calO(\nx).\label{Eq_sigmaGp_efficient}
\end{alignat}

\subsection{Posterior gradient evaluations for stationary kernels} \label{Sec_EfficientCalcPosterior__Grad}

The gradients of the posterior require the Jacobian $\pline{\kvec}{\xvec'}$, which could be calculated with \eqref{Eq_kern_der_r_ell_efficient}.
However, for stationary kernels it is more efficient to calculate the posterior gradients if the following matrix is used instead:
\begin{equation} \label{Eq_Hmat_r}
	\Hmatr
		= \left[\hvecr{1}, \ldots, \hvecr{\ndim} \right] 
		\in \mathbb{R}^{\nx \times \ndim} 
		\quad \calO(\nx \ndim),
\end{equation}
where Table~\ref{Tab_HmatCmat} provides $\hvecr{\ell}$ for different stationary kernels.
The gradient of the mean of the posterior is then calculated with
\begin{alignat}{2}
	\qvec_{\mu}
		&= \vvecres \odot \cvecr \quad
		&& \calO(\nx) \label{Eq_qvec_mu} \\
	\p{\muGP_f}{\xvec'}
		&= \hpsigK^{2} \, \qvec_{\mu}^\top \Hmatr \quad
		&&\calO(\nx \ndim), \label{Eq_Post_mean_grad_efficient}
\end{alignat}
where $\vvecres$ is from \eqref{Eq_vvecres} and the kernel-specific equations for $\cvecr$ are in Table~\ref{Tab_HmatCmat}.
The gradient of the variance of the posterior is calculated with
\begin{alignat}{2}
	\vveck
		&= \Lmat^{-\top} \vveckhalf \quad
		&& \calO(\nx^2) \label{Eq_vvec_kvec} \\
	\qvec_{\sigma}
		&= \vveck \odot \cvecr \quad
		&& \calO(\nx) \label{Eq_qvec_sigma} \\
	\p{\sigmaGP_f^2}{\xvec'}
		&= -2 \hpsigK^{4} \, \qvec_{\sigma}^\top \Hmatr \quad
		&& \calO(\nx \ndim), \label{Eq_Post_var_grad_efficient}
\end{alignat}
where $\left.\pline{k}{\xvec'} \right|_{(\xvec', \xvec')}$ was dropped since it is zero for stationary kernels.
In GpFlash the matrix $\Hmatr$ is never formed explicitly.
Rather, its entries are computed on the fly and applied directly to the required matrix-vector products.
Table~\ref{Tab_PosteriorComparison} compares how the posterior evaluation is implemented in GpFlash and other GP libraries.

\begin{table}[t]
\centering
\renewcommand{\arraystretch}{1.3}
\caption{Comparison of GP library methods to evaluate the posterior.
For GPy, GPyTorch, and GPflow the mean of the posterior is always evaluated with the variance, which has a computational cost of $\calO(\nx^2)$.
The Cholesky factor $\Lmat$ comes from \eqref{Eq_Lmat}, $\vvecres$ from \eqref{Eq_vvecres}, $\kvec$ from \eqref{Eq_kvec}, and $\vveckhalf$ from \eqref{Eq_vveckhalf}.}
\label{Tab_PosteriorComparison}
\adjustbox{max width=\linewidth}{%
\begin{tabular}{lcccccc}
 \toprule
 & scikit-learn & GPy & GPyTorch & GPflow & SMT & GpFlash \\
 \midrule
 Calculate $\Lmat$ \& $\vvecres$ once & \cmark & \cmark & \cmark & \cmark & \cmark & \cmark \\
 Mean-only evaluation & \cmark $\calO(\nx)$ & \xmark $\calO(\nx^2)$ & \xmark $\calO(\nx^2)$ & \xmark $\calO(\nx^2)$ & \cmark $\calO(\nx)$ & \cmark $\calO(\nx)$ \\
	Calculate $\muGP_f$ \& $\sigmaGP_f^2$ share $\kvec$ & \cmark & \cmark & \cmark & \cmark & \xmark & \cmark \\
 Gradient method & \xmark & Analytical & Autograd & Autograd & Analytical & Analytical \\
	$\pline{\muGP_f}{\xvec'}$ \& $\pline{\sigmaGP_f^2}{\xvec'}$ share $\pline{\kvec}{\xvec'}$ & \xmark & \cmark & \cmark & \cmark & \xmark & \cmark \\
 $\sigmaGP_f^2$ \& $\pline{\sigmaGP_f^2}{\xvec'}$ share $\vveckhalf$ & \xmark & \xmark & \cmark & \cmark & \xmark & \cmark \\
 \bottomrule
\end{tabular}%
}
\end{table}

\section{Benchmarking methodology} \label{Sec_Benchmarking}

\subsection{Hardware and software} \label{Sec_Benchmarking__CodeCompiler}

Numerical studies were performed on the Trillium supercomputer and on a MacBook Pro with an M3 Pro chip.
The Trillium nodes are AMD EPYC 9655 (Zen~5) processors with 24 core complex dies that each have 8 cores that share a common cache.
For each of the studies, one job was run per core complex die on Trillium and one in total on Mac so that jobs would not share cache.
On Mac, \texttt{caffeinate~-i} was used to prevent the computer from going to sleep during a run, and the \texttt{QOS\_CLASS\_USER\_INTERACTIVE} quality-of-service class was requested so that the timed calculations ran on a performance core rather than an efficiency core.

The code for all of the studies and GpFlash is written in C++, uses the matrix and linear algebra library Armadillo version 15.6.1, and makes use of 64-bit floating-point precision.
The compilers Apple Clang 21.0.0 and GCC 14 were used on Mac and Trillium, respectively, with the compilation flags listed in Table~\ref{Tab_build_config}.
On Mac, running macOS 27.0.1, Armadillo links to Apple's Accelerate framework for its BLAS and LAPACK methods, while on Trillium it links against AOCL 5.1.

\begin{table}[t]
\centering
\footnotesize
\setlength{\tabcolsep}{4pt}
\adjustbox{max width=\linewidth}{%
\begin{tabular}{@{} l p{10cm} @{}}
\hline
Flag & Effect \\
\hline
\texttt{-O3} & Aggressive optimization (vectorize, inline, unroll) \\
\texttt{-march=native} & Target the host instruction set \\
\texttt{-fno-math-errno} & Allows math functions such as \texttt{exp} to be vectorized \\
\texttt{-funroll-loops} & Unroll hot loops to expose instruction-level parallelism \\
\texttt{-fopenmp} & Enable OpenMP for the parallelized cases \\
\hline
\end{tabular}%
}
\caption{Compiler optimization flags, which are the same on Trillium and Mac.}
\label{Tab_build_config}
\end{table}

GpFlash is compared to the open source GP libraries scikit-learn version 1.9.1, GPy version 1.14.2, GPflow version 2.11.1, GPyTorch version 1.15.2, and SMT version 2.15.0.
All of these GP libraries used NumPy 2.4.2 and SciPy 1.17.1, except for GPflow, which requires a version of NumPy 1.x and thus used NumPy 1.26.4 with SciPy 1.17.1.
On Mac, NumPy 2.4.2 and SciPy 1.17.1 use the Accelerate framework for their BLAS and LAPACK methods, while NumPy 1.26.4 uses OpenBLAS.
On Trillium both versions of NumPy link against FlexiBLAS, which was configured to dispatch to AOCL.

\subsection{Testing methodology} \label{Sec_Benchmarking__TestingTiming}

Four threads were used for all timing calculations on both Trillium and Mac and for all of the GP libraries, while the memory studies were run with a single thread so that the peak memory is not perturbed by per-thread scratch buffers.
The studies were performed with $\ndim = 10$ and used the Mat\'ern $\frac{5}{2}$ kernel from \eqref{Eq_kern_Mat5f2}.

A fixed random seed was used for all of the studies to generate the input data.
Each memory and timing calculation was repeated 5 and 20 times, respectively, and the median wall time or memory was reported.
Cases are generally based on their predecessor with one or two changes to help isolate their performance differences.
The implementation used in GpFlash is always shown in green for each study.
The outputs of the different cases were compared to ensure that they produced the same result.

GpFlash uses Armadillo, which, like the BLAS and LAPACK libraries it calls, stores matrices in column-major order.
For that reason, it is more efficient to store the point locations in a matrix where each column is one point, \ie $\Xmat_\top = \Xmat^\top \in \mathbb{R}^{\ndim \times \nx}$, and likewise for $\Xmatdot_\top$, where the subscript makes it clear that the transpose operation is not being applied.

\section{Efficient training: the marginal log likelihood and its gradient} \label{Sec_Training}

\subsection{Study 1: Kernel matrix assembly} \label{Sec_Training__KernelMatrix}

This subsection considers different methods of forming the kernel matrix $\Kmat$ from \eqref{Eq_Kmat}.

\begin{description}[style=unboxed, leftmargin=0pt, labelindent=0pt, itemsep=2pt]
	\item[Case 1:] In scikit-learn and SMT the kernel matrix $\Kmat$ is calculated by leveraging its symmetry to mirror its entries but with no use of BLAS.
	The peak memory is $\nx^2$ and $\nx^2/2$ exponential evaluations are required.
	\item[Case 2:] GPyTorch and GPflow use the polarization identity from Section~\ref{Sec_EfficientKernelEval__BLAS} and the calculation $\Gmatdot = \Xmatdot_\top^\top \Xmatdot_\top$ from \eqref{Eq_Gmat} is done with the BLAS call \texttt{DGEMM}.
	The symmetry of $\Kmat$ is not leveraged, which thus requires $\nx^2$ exponential evaluations, and the peak storage is $2 \nx^2$.
	\item[Case 3:] The GPy implementation is similar to Case 2 but uses the BLAS method \texttt{DSYRK}, which only forms half of $\Gmatdot$.
	The lower triangular entries of $\Gmatdot$ are mirrored to the upper half of the matrix and the kernel is then evaluated for all $\nx^2$ entries.
	The peak memory is $2 \nx^2$.
	\item[Case 4:]
	Similar to Case 3 but the peak memory is reduced to $\nx^2$ by using a buffer matrix, $\Bmat$, to store the lower triangular entries of $\Gmatdot$ and then $\Kmat$, which only requires $\nx^2/2$ exponential evaluations.
	The GpFlash implementation uses the upper triangular entries of $\Bmat$ to store $\Cmatr$ from Table~\ref{Tab_HmatCmat}, which are used to calculate the gradient of $\ln(L)$ efficiently with \eqref{Eq_kern_der_gamma_ell_efficient}.
\end{description}

\begin{figure}[t]
\centering
\includegraphics[width=\textwidth]{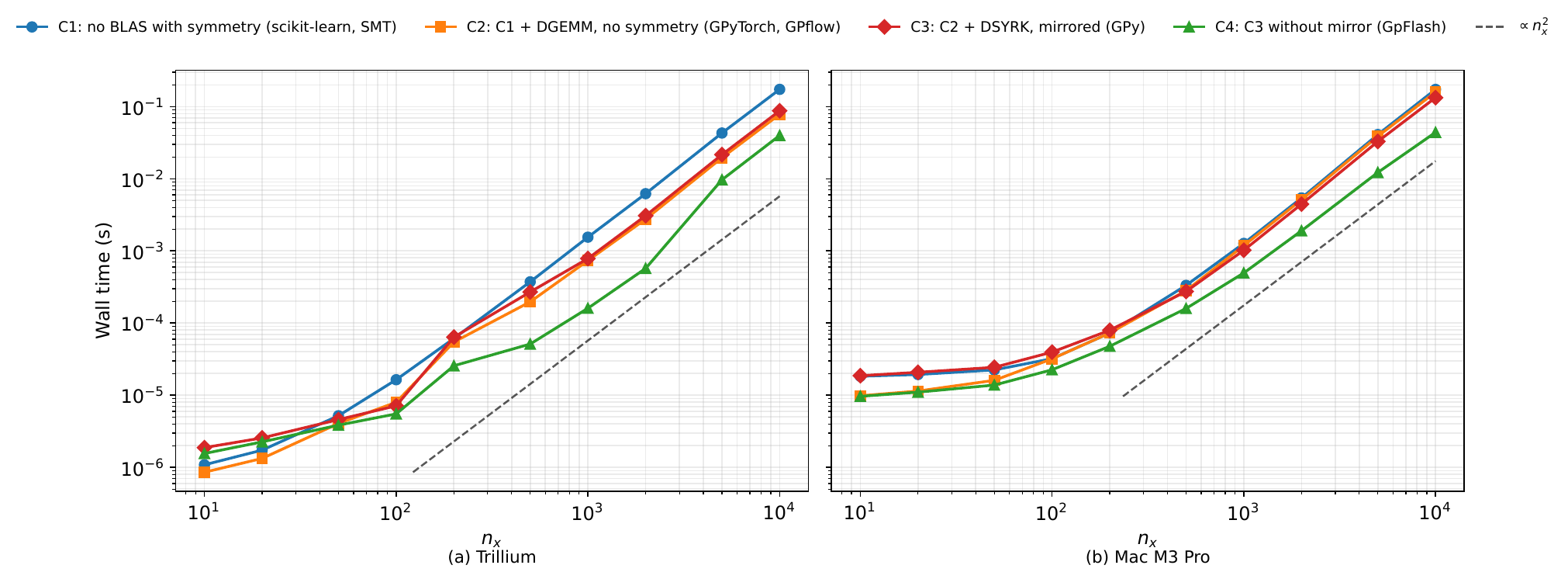}
\caption{Study 1: The median time to calculate the kernel matrix $\Kmat$ from \eqref{Eq_Kmat} using 20 runs, 4 threads, the Mat\'ern $\frac{5}{2}$ kernel, and $\ndim = 10$.}
\label{Fig_StudyTrain_KernelMatrix_10d_4threads}
\end{figure}

Figure~\ref{Fig_StudyTrain_KernelMatrix_10d_4threads} shows the median wall time of 20 runs using 4 threads for $\ndim = 10$ and $\nx$ varied from $10$ to $10^4$.
Case 4, which is used by GpFlash, is the fastest method for all cases with $\nx \geq 50$.
It is faster since it performs fewer computations and it also uses less memory.
Table~\ref{Tab_study_Kmat_4t} indicates the speedup on Trillium and Mac that each case provides relative to the previous one.

\begin{table}[t]
\centering\small
\adjustbox{max width=\linewidth}{%
\begin{tabular}{l|ccc}
\hline
 & $n_x = 10^{2}$ & $n_x = 10^{3}$ & $n_x = 10^{4}$ \\
\hline
C1: no BLAS with symmetry (scikit-learn, SMT) & baseline & baseline & baseline \\
C2: C1 + \texttt{DGEMM}, no symmetry (GPyTorch, GPflow) & \green{2.1$\times$} / 1.0$\times$ & \green{2.1$\times$} / 1.1$\times$ & \green{2.2$\times$} / 1.1$\times$ \\
C3: C2 + \texttt{DSYRK}, mirrored (GPy) & 1.1$\times$ / \red{0.8$\times$} & \red{0.9$\times$} / 1.2$\times$ & \red{0.9$\times$} / 1.2$\times$ \\
C4: C3 without mirror (GpFlash) & 1.3$\times$ / 1.8$\times$ & \green{4.9$\times$} / \green{2.1$\times$} & \green{2.2$\times$} / \green{3.0$\times$} \\
\hline
final vs.\ baseline & \green{3.0$\times$} / 1.4$\times$ & \green{9.7$\times$} / \green{2.6$\times$} & \green{4.4$\times$} / \green{3.9$\times$} \\
\hline
\end{tabular}
}

\caption{
Study 1: Table entries indicate the speedup that each case provides on Trillium / Mac relative to its predecessor to construct the kernel matrix $\Kmat$.
Individual times for the cases can be seen in Figure~\ref{Fig_StudyTrain_KernelMatrix_10d_4threads}.
\red{Red} $<1$; \green{green} $\geq 2\times$.}
\label{Tab_study_Kmat_4t}
\end{table}

\subsection{Study 2: Calculating $\Psimat$} \label{Sec_Training__PsiMatrix}

This subsection explores different methods of calculating $\Psimat$ from \eqref{Eq_Psimat}, which is needed to calculate the gradient of $\ln(L)$ efficiently with \eqref{Eq_gradient_ln_L_with_frobenius_Psimat}.
SMT is the only GP library considered in this paper that does not form $\Psimat$.
Instead, it uses \eqref{Eq_grad_ln_lkd_SMT} to calculate the gradient of $\ln(L)$.
The other GP library whose method is not amongst the cases for this study is GPyTorch, which forms $\Psimat$ through the automatic differentiation of its Cholesky factorization rather than by explicitly inverting $\Sigma$.
For all of the cases in this study, the Cholesky factor $\Lmat$ from \eqref{Eq_Lmat} is calculated with the LAPACK method \texttt{DPOTRF}.

\begin{description}[style=unboxed, leftmargin=0pt, labelindent=0pt, itemsep=2pt]
	\item[Case 1:] The BLAS method \texttt{DTRSM} is used to calculate $\Sigma^{-1}$ using $\Lmat$ by applying two triangular solves to the identity matrix.
	This is the method used by both scikit-learn and GPflow.
	The matrix $\Psimat$ is then formed resulting in a peak memory of $2\nx^2$.
	\item[Case 2:] The GPy method is similar to Case 1 but the LAPACK method \texttt{DPOTRI} is used to calculate the lower triangular entries of $\Sigma^{-1}$ using $\Lmat$.
	The upper half entries of $\Sigma^{-1}$ are formed by mirroring its lower entries and then $\Psimat$ is formed as a new matrix, resulting in a peak memory of $2\nx^2$.
	\item[Case 3:] Case 2 but only the lower half entries of $\Sigma^{-1}$ are formed.
	The matrix $\Psimat$ is formed by overwriting the entries of $\Sigma^{-1}$ in place, reducing the computational cost and the peak memory to $\nx^2$.
	\item[Case 4:] The GpFlash implementation has a peak memory of $\nx^2$ but the LAPACK method \texttt{DPOTRI} is not applied on the full matrix.
	Instead, its constituent routines \texttt{DTRTRI} and \texttt{DLAUUM} are applied recursively to diagonal blocks no larger than $128 \times 128$.
	These blocks are then combined using the BLAS methods \texttt{DTRMM} and \texttt{DSYRK}.
	This requires the same number of operations as Case 3 and its wall time is within 10\% of that of Case 3 on Trillium, but it was found to be up to 2.3 times faster on Mac.
\end{description}

A reverse-Cholesky method developed by Smith~\cite{smith_differentiation_1995}, which was popularized in the GP literature by Toal et al.~\cite{toal_adjoint_2009}, was also considered.
This method constructs $\Psimat$ directly from $\Lmat$ without ever forming $\Sigma^{-1}$ explicitly.
The peak memory is $\nx^2$ and it requires fewer calculations than forming $\Sigma^{-1}$ then $\Psimat$.
However, this method does not leverage BLAS or LAPACK methods and was found to be significantly slower than the other cases considered.

\begin{figure}[t]
\centering
\includegraphics[width=\textwidth,height=0.377\textwidth,keepaspectratio]{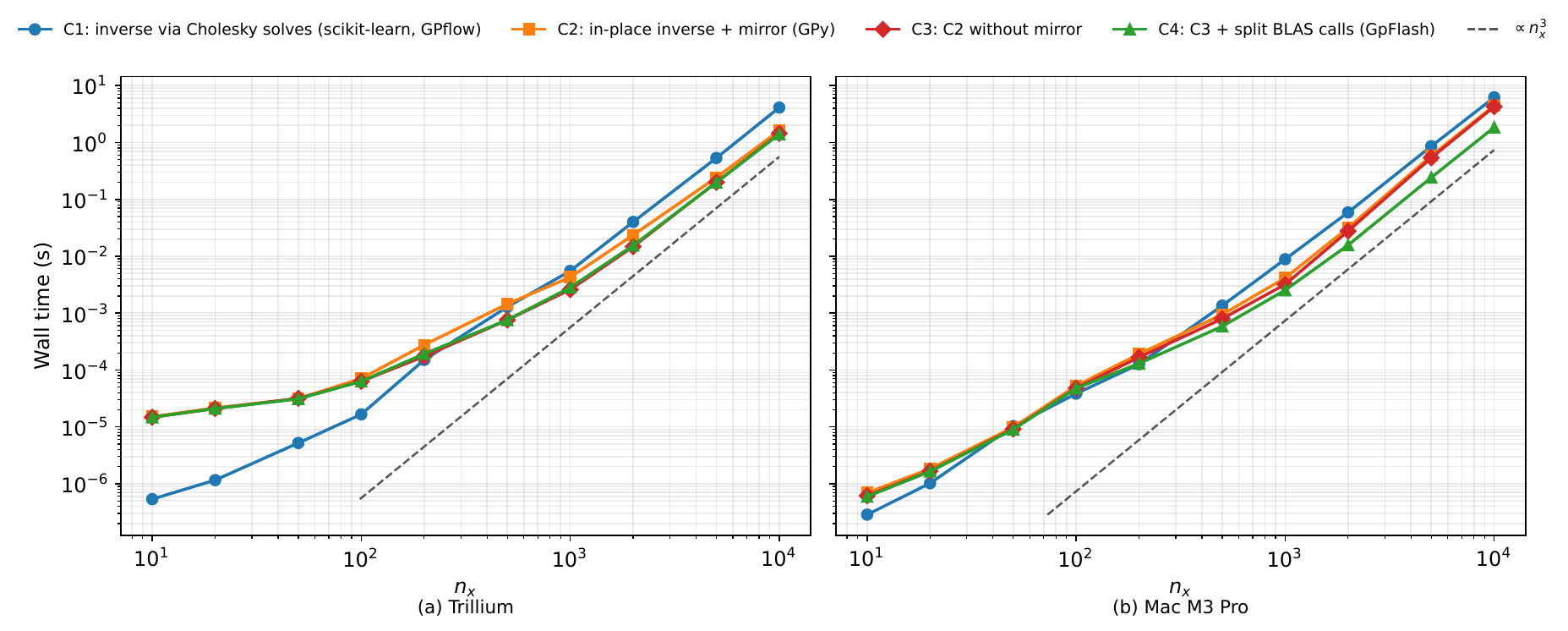}
\caption{Study 2: Median wall time of different methods to form the adjoint matrix $\Psimat$.
For each case, 20 runs with $\ndim = 10$ are used along with 4 threads and the Mat\'ern $\frac{5}{2}$ kernel.}
\label{Fig_StudyTrain_AdjointMatrix_10d_4threads}
\end{figure}

Figure~\ref{Fig_StudyTrain_AdjointMatrix_10d_4threads} and Table~\ref{Tab_study_Psimat_4t} compare the median wall time for the different cases using 4 threads and for data with $\ndim = 10$.
Case 4 is the one used by GpFlash since it is the fastest on Mac for $\nx \geq 500$ and within 10\% of Case 3 on Trillium.

\begin{table}[t]
\centering\small
\adjustbox{max width=\linewidth}{%
\begin{tabular}{l|ccc}
\hline
 & $n_x = 10^{2}$ & $n_x = 10^{3}$ & $n_x = 10^{4}$ \\
\hline
C1: inverse via Cholesky solves (scikit-learn, GPflow) & baseline & baseline & baseline \\
C2: in-place inverse + mirror (GPy) & \red{0.2$\times$} / \red{0.7$\times$} & 1.3$\times$ / \green{2.1$\times$} & \green{2.5$\times$} / 1.4$\times$ \\
C3: C2 without mirror & 1.1$\times$ / 1.1$\times$ & 1.6$\times$ / 1.3$\times$ & 1.1$\times$ / 1.0$\times$ \\
C4: C3 + split BLAS calls (GpFlash) & 1.0$\times$ / 1.0$\times$ & \red{0.9$\times$} / 1.3$\times$ & 1.0$\times$ / \green{2.3$\times$} \\
\hline
final vs.\ baseline & \red{0.3$\times$} / \red{0.8$\times$} & 2.0$\times$ / \green{3.5$\times$} & \green{2.9$\times$} / \green{3.4$\times$} \\
\hline
\end{tabular}
}

\caption{Study 2: Speedup provided by each case relative to its predecessor to form $\Psimat$.
Each table entry indicates the speedup on Trillium / Mac for data that has 10 dimensions and with the median time calculated using 4 threads and 20 runs.
\red{Red} $<1$; \green{green} $\geq 2\times$.}
\label{Tab_study_Psimat_4t}
\end{table}

\subsection{Study 3: Gradient for the marginal log likelihood} \label{Sec_Training__MllGradient}

In this subsection different methods of calculating the analytical gradient of the marginal log likelihood with respect to $\hpgammavec$ are presented.
The derivative $\pline{\eta}{\hpgamma_\ell}$ from \eqref{Eq_der_eta_wrt_gamma} is omitted from all of the cases since it is not used by any of the other GP libraries and its cost is small relative to the total gradient cost.
The automatic differentiation used by GPyTorch and GPflow is not considered since Armadillo does not support automatic differentiation.

\begin{figure}[t]
\centering
\begin{subfigure}[t]{0.46\textwidth}
\centering
\begin{tikzpicture}[scale=0.5, line join=round]
 \foreach \i in {1,...,6}{ \fill[black!12] (\i-1,1-\i) rectangle (\i,-\i); }
 \foreach \i in {2,...,6}{ \fill[cyan!25] (0,1-\i) rectangle (1,-\i); }
 \foreach \j in {2,...,6}{ \fill[cyan!25] (\j-1,0) rectangle (\j,-1); }
 \foreach \i in {3,...,6}{ \fill[green!28] (1,1-\i) rectangle (2,-\i); }
 \foreach \j in {3,...,6}{ \fill[green!28] (\j-1,-1) rectangle (\j,-2); }
 \draw[step=1,black!45] (0,0) grid (6,-6);
 \draw[black!60,thick] (0,0) rectangle (6,-6);
 \draw[-{Latex[length=1.6mm]},thick,blue!65!black] (0.5,-1.5) -- (0.5,-5.5);
 \draw[-{Latex[length=1.6mm]},thick,blue!65!black] (1.5,-0.5) -- (5.5,-0.5);
 \draw[-{Latex[length=1.6mm]},thick,green!50!black] (1.5,-2.5) -- (1.5,-5.5);
 \draw[-{Latex[length=1.6mm]},thick,green!50!black] (2.5,-1.5) -- (5.5,-1.5);
\end{tikzpicture}
\caption{Transpose layout}
\label{Fig_TriangularStorage__mirrored}
\end{subfigure}
\hfill
\begin{subfigure}[t]{0.46\textwidth}
\centering
\begin{tikzpicture}[scale=0.5, line join=round]
 \foreach \i in {1,...,6}{ \fill[black!12] (\i-1,1-\i) rectangle (\i,-\i); }
 \foreach \i in {2,...,6}{ \fill[cyan!25] (0,1-\i) rectangle (1,-\i); }
 \foreach \i in {1,...,5}{ \fill[cyan!25] (5,1-\i) rectangle (6,-\i); }
 \foreach \i in {3,...,6}{ \fill[green!28] (1,1-\i) rectangle (2,-\i); }
 \foreach \i in {1,...,4}{ \fill[green!28] (4,1-\i) rectangle (5,-\i); }
 \draw[step=1,black!45] (0,0) grid (6,-6);
 \draw[black!60,thick] (0,0) rectangle (6,-6);
 \draw[-{Latex[length=1.6mm]},thick,blue!65!black] (0.5,-1.5) -- (0.5,-5.5);
 \draw[-{Latex[length=1.6mm]},thick,blue!65!black] (5.5,-0.5) -- (5.5,-4.5);
 \draw[-{Latex[length=1.6mm]},thick,green!50!black] (1.5,-2.5) -- (1.5,-5.5);
 \draw[-{Latex[length=1.6mm]},thick,green!50!black] (4.5,-0.5) -- (4.5,-3.5);
\end{tikzpicture}
\caption{Column-major layout}
\label{Fig_TriangularStorage__colmajor}
\end{subfigure}
\caption{A matrix can be used to store two symmetric matrices (only one diagonal needs to be stored for GpFlash), split across its lower and upper triangular entries.
The layout in (a) is the traditional method to store a symmetric matrix in the upper triangular entries.
GpFlash uses the layout in (b) since it is more efficient with its column-major memory.}
\label{Fig_TriangularStorage}
\end{figure}
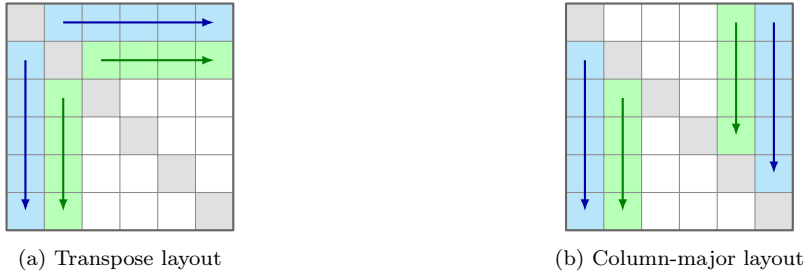

\begin{description}[style=unboxed, leftmargin=0pt, labelindent=0pt, itemsep=2pt]
	\item[Case 1:] In SMT the gradient of $\ln(L)$ is calculated with \eqref{Eq_grad_ln_lkd_SMT}.
	This requires forming the matrix $\pline{\Sigma}{\hpgamma_\ell}$ for each of the $\ndim$ entries of the gradient vector and performing two triangular solves with the Cholesky factor.
	The computational cost scales as $\calO(\nx^3 \ndim)$ and this implementation has a peak memory of $5\nx^2$ to store $\Lmat$, $\Cmatr$, $\pline{\Sigma}{\hpgamma_\ell}$, and the results of the two triangular solves.
	SMT requires even more memory since it also stores the componentwise distances for all pairs of points, increasing the peak memory by $\nx^2 \ndim / 2$.
	\item[Case 2:] In scikit-learn $\Psimat$ is formed along with all $\ndim$ matrices $\pline{\Kmat}{\hpgamma_\ell} = \Hmatgam{\ell} \odot \Cmatr$ from \eqref{Eq_kern_der_gamma_ell_efficient} without making use of symmetry before then calculating the Frobenius inner products with \eqref{Eq_gradient_ln_L_with_frobenius_Psimat}.
	This results in a peak memory of $\nx^2 (\ndim + 2)$ to store $\Sigma^{-1}$, $\Psimat$, along with the $\ndim$ derivative matrices, and it requires $\nx^2$ exponential evaluations in addition to those needed to form $\Kmat$.
	\item[Case 3:] GPy's cython path first forms the matrix $\Amat = \Psimat \odot \Cmatr$ and then for each of the $\ndim$ dimensions it calculates the Frobenius inner product with \eqref{Eq_gradient_ln_L_with_frobenius_Amat} without forming the full matrix $\Hmatgam{\ell}$.
	Like Case 2, this requires $\nx^2$ exponential evaluations beyond those used to form $\Kmat$, but it uses fewer multiplications since $\Cmatr$ is applied to $\Psimat$ rather than to $\ndim$ derivative matrices.
	The peak memory is $4\nx^2$ to store $\Sigma^{-1}$, $\Psimat$, $\Cmatr$, and $\Amat$.
	\item[Case 4:] The symmetry of $\Amat$ and $\Hmatgam{\ell}$ is exploited to halve both the number of Frobenius inner products and the exponential evaluations needed to calculate the entries of $\Cmatr$.
	The peak memory is reduced to $\nx^2$ by only storing the lower half of $\Sigma^{-1}$ and doing all calculations on the fly rather than forming the additional matrices explicitly.
	\item[Case 5:] The entries of $\Cmatr$, which were calculated with \eqref{Eq_c_rdot} when forming $\Kmat$, are stored in the upper half of the buffer matrix $\Bmat$ with the memory layout shown in Figure~\ref{Fig_TriangularStorage__mirrored}.
	No additional exponential evaluations are required and the peak memory remains $\nx^2$
	since $\Sigma^{-1}$ is stored in the lower half of $\Bmat$.
	\item[Case 6:] Same process as Case 5 but the layout of the data in the upper triangular entries of $\Bmat$ is now column-major as shown in Figure~\ref{Fig_TriangularStorage__colmajor}.
	This is the implementation used in GpFlash and its peak memory is $\nx^2$.
\end{description}

\begin{figure}[t]
\centering
\includegraphics[width=\textwidth]{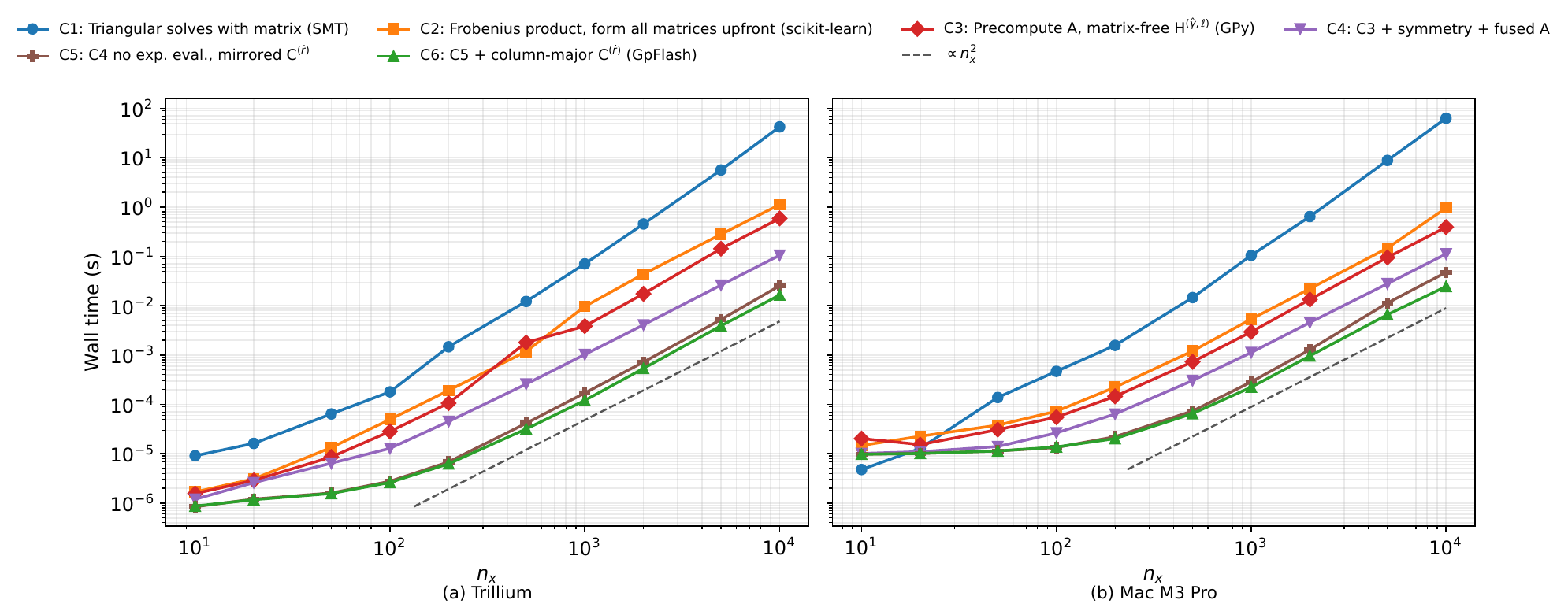}
\caption{Study 3: The median wall time to calculate the gradient of $\ln(L)$ using 20 runs, 4 threads, and $\ndim = 10$ for the Mat\'ern $\frac{5}{2}$ kernel.}
\label{Fig_StudyTrain_MllGradient_10d_4threads}
\end{figure}

The three orders of magnitude speedup provided at $\nx = 10^4$ from Case 1 to 6 seen in Figure~\ref{Fig_StudyTrain_MllGradient_10d_4threads} and Table~\ref{Tab_study_mll_grad_4t} is the result of more efficient computations and memory access.
Case 6, which is used by GpFlash, requires the least amount of memory relative to the implementations in the GP libraries considered at only $\nx^2$.

\begin{table}[t]
\centering\small
\adjustbox{max width=\linewidth}{%
\begin{tabular}{l|ccc}
\hline
 & $n_x = 10^{2}$ & $n_x = 10^{3}$ & $n_x = 10^{4}$ \\
\hline
C1: Triangular solves with matrix (SMT) & baseline & baseline & baseline \\
C2: Frobenius product, form all matrices upfront (scikit-learn) & \green{3.6$\times$} / \green{6.5$\times$} & \green{7.3$\times$} / \green{19.6$\times$} & \green{37.6$\times$} / \green{65.9$\times$} \\
C3: Precompute $\Amat$, matrix-free $\Hmatgam{\ell}$ (GPy) & 1.7$\times$ / 1.3$\times$ & \green{2.5$\times$} / 1.8$\times$ & 1.9$\times$ / \green{2.4$\times$} \\
C4: C3 + symmetry + fused $\Amat$ & \green{2.2$\times$} / \green{2.1$\times$} & \green{3.8$\times$} / \green{2.6$\times$} & \green{5.6$\times$} / \green{3.5$\times$} \\
C5: C4 no exp.\ eval., mirrored $\Cmat^{(\rdot)}$ & \green{4.7$\times$} / 2.0$\times$ & \green{6.0$\times$} / \green{4.0$\times$} & \green{4.1$\times$} / \green{2.4$\times$} \\
C6: C5 + column-major $\Cmat^{(\rdot)}$ (GpFlash) & 1.1$\times$ / 1.0$\times$ & 1.4$\times$ / 1.3$\times$ & 1.5$\times$ / 1.9$\times$ \\
\hline
final vs.\ baseline & \green{68.7$\times$} / \green{34.2$\times$} & \green{586$\times$} / \green{468$\times$} & \green{2560$\times$} / \green{2561$\times$} \\
\hline
\end{tabular}
}

\caption{Study 3: Each table entry shows the speedup on Trillium / Mac provided by each case relative to its predecessor to calculate the gradient of $\ln(L)$ when using $\ndim = 10$, 4 threads, and the Mat\'ern $\frac{5}{2}$ kernel.
\red{Red} $<1$; \green{green} $\geq 2\times$.}
\label{Tab_study_mll_grad_4t}
\end{table}

\subsection{Complete algorithm for GpFlash} \label{Sec_Training__GpFlashAlgorithm}

Algorithm~\ref{Alg_FullCalc_GpFlash} details how GpFlash calculates the marginal log likelihood $\ln(L)$ and its gradient.
A unique feature of GpFlash is that it only requires one matrix, $\Bmat \in \mathbb{R}^{\nx \times \nx}$, to be stored.
This is achieved by reusing the lower and upper triangular entries in $\Bmat$ to store two different symmetric matrices.

The GP libraries scikit-learn and GPy evaluate the gradient of $\ln(L)$ using a factored form similar to the one presented in Section~\ref{Sec_EfficientKernelEval__CommonTerms} to have fewer exponential evaluations.
GpFlash requires only $\nx^2/2$ exponential evaluations to evaluate $\ln(L)$ and its gradient, which is 3 and 4 times fewer than scikit-learn and GPy, respectively.
GpFlash achieves this by leveraging the symmetry of the kernels and by storing the entries of $\Cmatr$ (equations in Table~\ref{Tab_HmatCmat}) in the upper half of $\Bmat$ after they are used to calculate $\Kmat$.
$\Cmatr$ is then used for the gradient calculation using \eqref{Eq_kern_der_gamma_ell_efficient}.

\begin{algorithm}[t!]
\caption{
GpFlash implementation to calculate $\ln(L)$ and its gradient when using a stationary kernel.
A single buffer matrix is used: $\Bmat \in \mathbb{R}^{\nx \times \nx}$.
The lower half of the buffer matrix along with its diagonal is denoted $\Bmat^{(l)}$, which holds $\Kmat$, $\Sigma$, $\Lmat$, and then finally $\Sigma^{-1}$.
The upper half of the buffer matrix, $\Bmat^{(u)}$, holds $\Cmatr$ using the layout shown in Figure~\ref{Fig_TriangularStorage__colmajor}.}
\label{Alg_FullCalc_GpFlash}
\begin{algorithmic}[1]
\Require $\Xmat_\top \in \mathbb{R}^{\ndim \times \nx}$, $\fvectil \in \mathbb{R}^{\nx}$, $\hpgammavec \in \mathbb{R}_+^{\ndim}$, $\hpsigK^2 \geq 0$, $\hpstdfval^{\,2} \geq 0$
\Ensure $\ln(L)$ and its gradient $\pline{\ln(L)}{\hpgammavec} \in \mathbb{R}^{\ndim}$

\For{$(i, j)$ with $1 \le j < i \le \nx$}
 \State $\centk_{ij} \leftarrow \centk(\xvec_i, \xvec_j, \hpgammavec)$ \Comment{Eq.\ for kernels in Table~\ref{Tab_HmatCmat}}
 \State $b^{(l)}_{ij} \leftarrow \sentk_{ij}\,\centk_{ij}$ \Comment{$\Bmat^{(l)}$ stores $\Kmat$; Eq.~\eqref{Eq_kern_efficient}}
 \State $b^{(u)}_{ij} \leftarrow \sentr_{ij}\,\centk_{ij}$ \Comment{$\Bmat^{(u)}$ stores $\Cmatr$}
\EndFor
\State $\eta \leftarrow \bigl(1 + \textstyle\max_i \sum_{j \neq i} \lvert k_{ij} \rvert\bigr) / (\condmax - 1)$ \Comment{Eq.~\eqref{Eq_eta}}
\State $b^{(l)}_{ij} \leftarrow \hpsigK^{2}\,b^{(l)}_{ij}$ for $i > j$ \Comment{Off-diagonal $\Sigma$; Eq.~\eqref{Eq_Sigma}}
\State $b^{(l)}_{ii} \leftarrow \hpsigK^{2}\,(1 + \eta) + \hpstdfval^{\,2}$ for $i = 1, \ldots, \nx$ \Comment{Diagonal $\Sigma_{ii}$; Eq.~\eqref{Eq_Sigma}}
\State $\Bmat^{(l)} \leftarrow \texttt{DPOTRF}(\Bmat^{(l)})$ \Comment{$\Bmat^{(l)}$ stores $\Lmat$; Eq.~\eqref{Eq_Lmat}}
\State $\vveconehalf \leftarrow \Lmat^{-1} \one$ \Comment{Eq.~\eqref{Eq_vveconehalf}}
\State $\vvecftilhalf \leftarrow \Lmat^{-1} \fvectil$ \Comment{Eq.~\eqref{Eq_vvecftilhalf}}
\State $\hpbeta \leftarrow \langle \vvecftilhalf, \, \vveconehalf \rangle / \| \vveconehalf \|^{2}$ \Comment{Eq.~\eqref{Eq_hpbeta_efficient}}
\State $\vvecreshalf \leftarrow \vvecftilhalf - \hpbeta\, \vveconehalf$ \Comment{Eq.~\eqref{Eq_half_res_solve}}
\State $\ln(L) \leftarrow -\textstyle\sum_{i} \ln b^{(l)}_{ii} - \tfrac{1}{2} \| \vvecreshalf \|^2$ \Comment{Eq.~\eqref{Eq_ln_lkd_noisy_efficient}}
\State $\vvecres \leftarrow \Lmat^{-\top} \vvecreshalf$ \Comment{Eq.~\eqref{Eq_vvecres}}
\State $\Bmat^{(l)} \leftarrow \Sigma^{-1}$ \Comment{Case~4 of Sec.~\ref{Sec_Training__PsiMatrix}}
\For{$(i, j)$ with $1 \le j < i \le \nx$}
 \State $a_{ij} \leftarrow \tfrac{1}{2}((\vres)_i (\vres)_j - b^{(l)}_{ij}) \cdot b^{(u)}_{ij}$ \Comment{$\Amat_{ij} = (\Psimat \odot \Cmatr)_{ij}$; Eq.~\eqref{Eq_Amat}}
\EndFor
\State $\zeta \leftarrow \tfrac{1}{2}\textstyle\sum_{i} \bigl( (\vres)_i^2 - b^{(l)}_{ii} \bigr)$ \Comment{$\Wmat = \Idty$; Eq.~\eqref{Eq_zeta}}
\State $\pline{\eta}{\hpgamma_\ell} \leftarrow \bigl( \textstyle\sum_{j \neq i^\star} \sgn(k_{i^\star j})\, \hentgam{\ell}_{i^\star j}\, b^{(u)}_{i^\star j} \bigr) / (\condmax - 1)$ $\forall \, \ell \leq \ndim$ \Comment{Eq.~\eqref{Eq_der_eta_wrt_gamma}}
\State $\pline{\ln(L)}{\hpgamma_\ell} \leftarrow \hpsigK^{2}\bigl( 2\textstyle\sum_{i>j} a_{ij}\, \hentgam{\ell}_{ij} + \zeta\, \pline{\eta}{\hpgamma_\ell} \bigr)$ $\forall \, \ell \leq \ndim$ \Comment{Eq.~\eqref{Eq_gradient_ln_L_with_frobenius_Amat}}
\State \Return $\ln(L)$, $\pline{\ln(L)}{\hpgammavec}$
\end{algorithmic}
\end{algorithm}

\section{Benchmarking GpFlash for training} \label{Sec_BenchmarkTraining}

\subsection{Training costs} \label{Sec_BenchmarkTraining__GpLibTrainingComp}

In this subsection the cost of evaluating the marginal log likelihood and its gradient is compared for different GP libraries.
For $\nx > 800$, GPyTorch uses an approximate method to calculate $\ln(L)$ and its gradient, which is faster but results in substantial errors that were found to exceed 20\% in spot testing.
To have a fair comparison for all of the GP libraries, the settings for GPyTorch were changed so that the exact solver was used in all cases.

\begin{figure}[t]
\centering
\includegraphics[width=\textwidth]{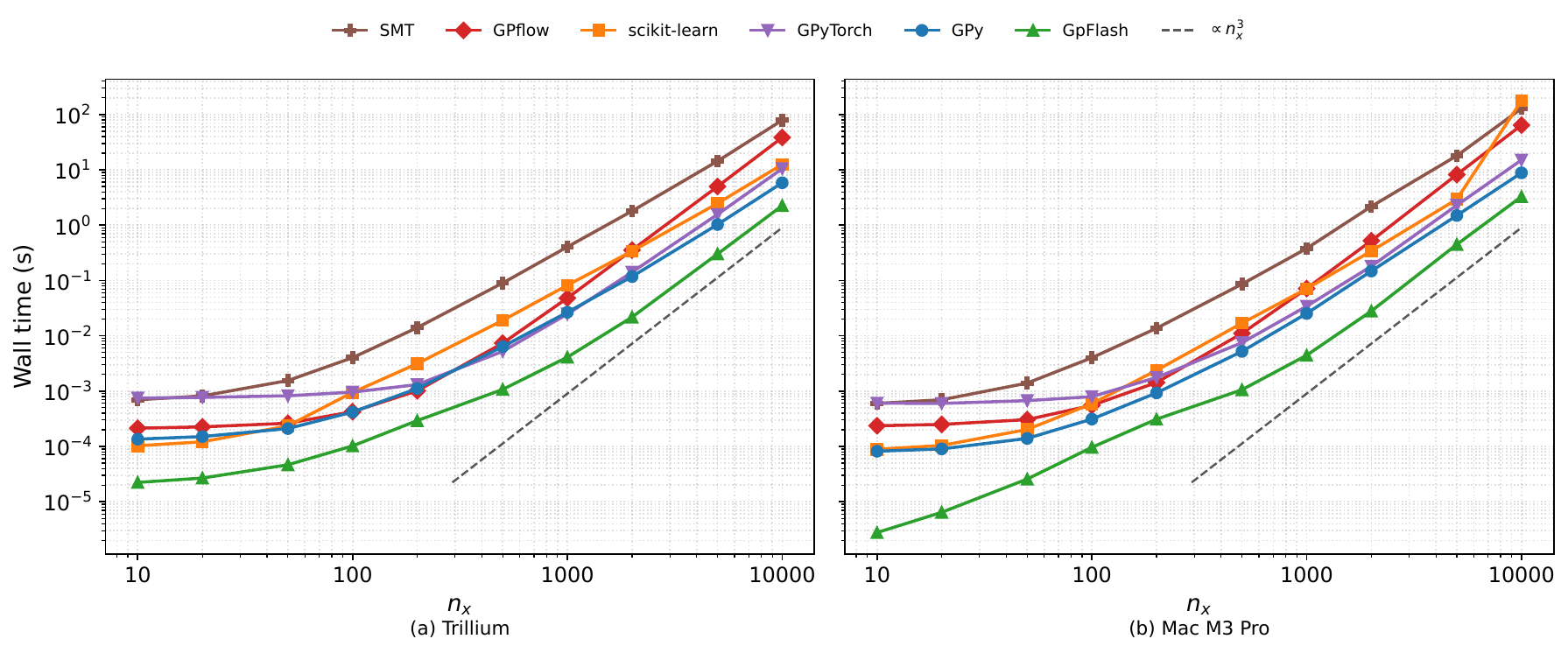}
\caption{
	Median wall time for the GP libraries to calculate $\ln(L)$ and its gradient for the Mat\'ern $\frac{5}{2}$ kernel at $\ndim = 10$ using 4 threads, on (a) Trillium and (b) Mac.}
\label{Fig_StudyTrain_GpLibsComparison_10d_4threads}
\end{figure}

Figure~\ref{Fig_StudyTrain_GpLibsComparison_10d_4threads} compares the wall time of the GP libraries to calculate the marginal log likelihood $\ln(L)$ and its gradient for problems with 10 dimensions.
Table~\ref{Tab_speedup_gp_libs_train_4t} indicates that GpFlash is 2.6 to 99 times faster than the other GP libraries.
The slowest GP library is generally SMT, which calculates the gradient of $\ln(L)$ using the computationally expensive method from \eqref{Eq_grad_ln_lkd_SMT}.

\begin{table}[t]
\centering\footnotesize
\setlength{\tabcolsep}{4pt}
\adjustbox{max width=\linewidth}{%
\begin{tabular}{l|ccc}
\hline
 & $n_x = 10^{2}$ & $n_x = 10^{3}$ & $n_x = 10^{4}$ \\
\hline
SMT        & \green{39.4$\times$} / \green{41.7$\times$}
           & \green{98.8$\times$} / \green{85.4$\times$}
           & \green{35.0$\times$} / \green{39.1$\times$} \\
GPflow     & \green{4.2$\times$} / \green{5.7$\times$}
           & \green{11.7$\times$} / \green{16.1$\times$}
           & \green{17.0$\times$} / \green{19.6$\times$} \\
scikit-learn& \green{9.3$\times$} / \green{6.1$\times$}
           & \green{20.0$\times$} / \green{16.0$\times$}
           & \green{5.5$\times$} / \green{54.0$\times$} \\
GPyTorch   & \green{9.4$\times$} / \green{8.3$\times$}
           & \green{6.0$\times$} / \green{7.7$\times$}
           & \green{4.6$\times$} / \green{4.6$\times$} \\
GPy        & \green{4.1$\times$} / \green{3.2$\times$}
           & \green{6.5$\times$} / \green{5.7$\times$}
           & \green{2.6$\times$} / \green{2.7$\times$} \\
\hline
Min speedup& \green{4.1$\times$} / \green{3.2$\times$}
           & \green{6.0$\times$} / \green{5.7$\times$}
           & \green{2.6$\times$} / \green{2.7$\times$} \\
Max speedup& \green{39.4$\times$} / \green{41.7$\times$}
           & \green{98.8$\times$} / \green{85.4$\times$}
           & \green{35.0$\times$} / \green{54.0$\times$} \\
\hline
\end{tabular}%
}

\caption{Table entries indicate how much faster GpFlash is on Trillium / Mac relative to other GP libraries using 4 threads, the Mat\'ern $\frac{5}{2}$ kernel, and with $\ndim = 10$.
\green{Green} $\geq 2\times$.}
\label{Tab_speedup_gp_libs_train_4t}
\end{table}

\begin{figure}[t]
\centering
\includegraphics[width=\textwidth]{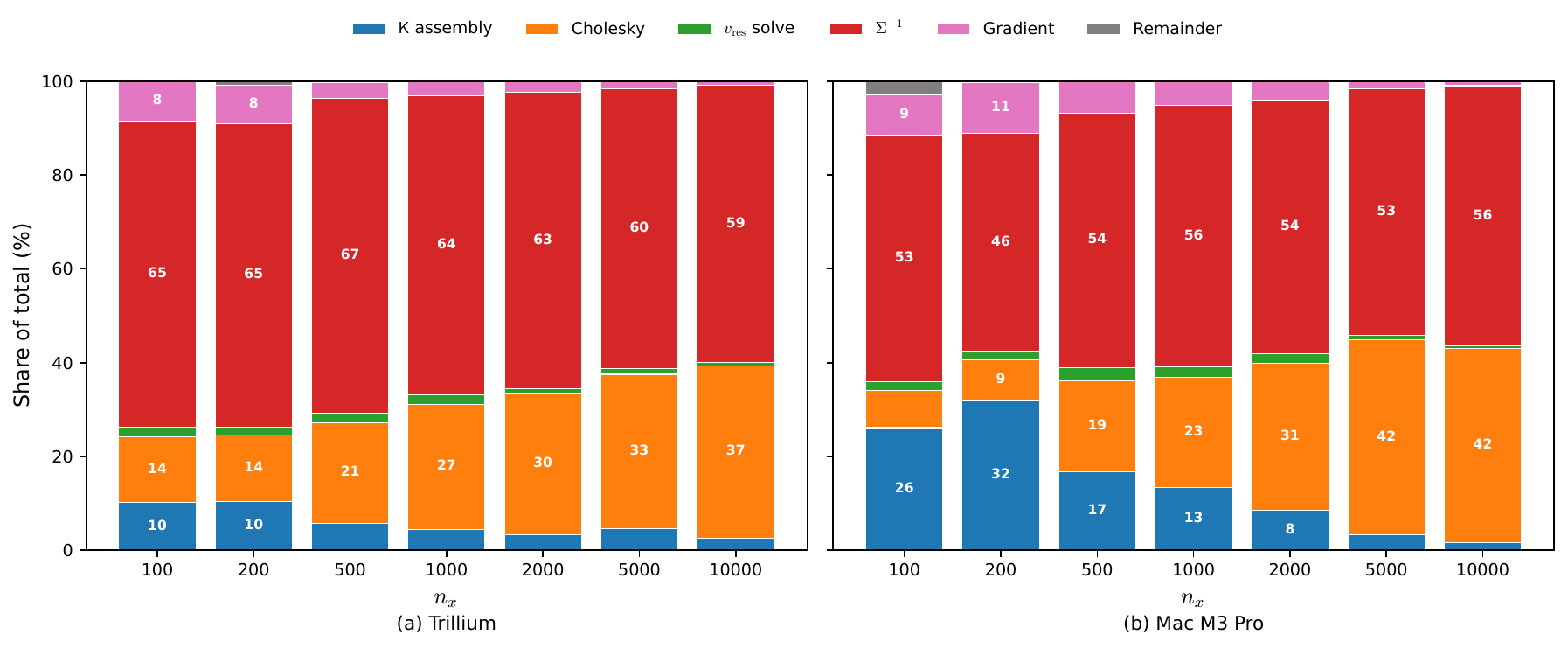}
\caption{The median time to calculate each step to form $\ln(L)$ and its gradient with GpFlash is shown for 10-dimensional problems, using 4 threads, and the Mat\'ern $\frac{5}{2}$ kernel.
}
\label{Fig_StudyTrain_GpFlashBreakdown_10d_4threads}
\end{figure}

Figure~\ref{Fig_StudyTrain_GpFlashBreakdown_10d_4threads} breaks down the cost for each step to calculate the marginal log likelihood and its gradient.
For the largest case with $\nx = 10^4$, the BLAS and LAPACK operations used to calculate $\Lmat$ and $\Sigma^{-1}$ take $96\%$ of the wall time on Trillium and $97\%$ on Mac.
Little further algorithmic optimization is therefore possible for large $\nx$, and the wall time is nearly independent of $\ndim$, since the cost of computing $\Lmat$ and $\Sigma^{-1}$ does not depend on it.

\subsection{Memory usage during training} \label{Sec_BenchmarkTraining__GpLibMemoryComp}

\begin{figure}[t]
\centering
\includegraphics[width=\textwidth]{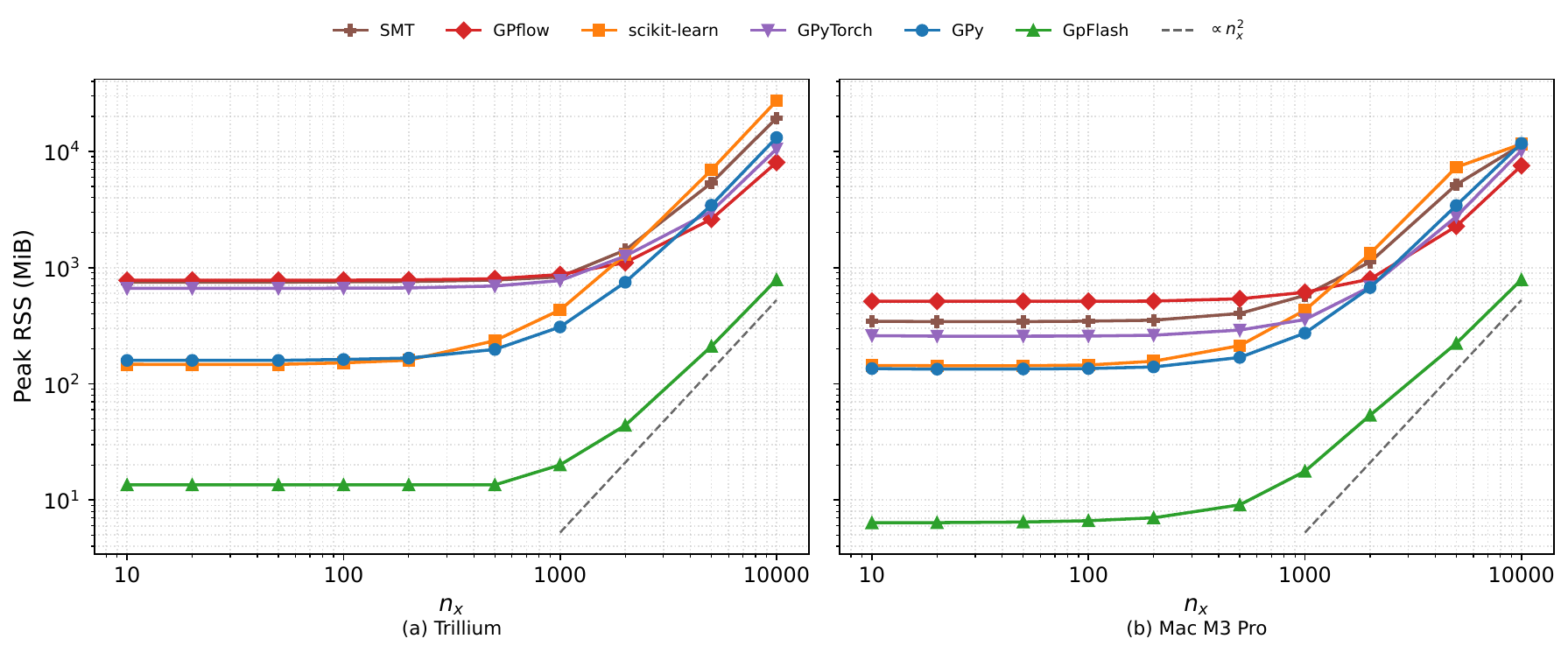}
\caption{Median peak memory measured across 5 runs for the calculation of $\ln(L)$ and of its gradient with $\ndim = 10$, a single thread, and the Mat\'ern $\frac{5}{2}$ kernel.}
\label{Fig_StudyTrain_GpLibsMemory_10d_1thread}
\end{figure}

Figure~\ref{Fig_StudyTrain_GpLibsMemory_10d_1thread} shows the peak memory used by GpFlash and the other GP libraries to calculate $\ln(L)$ and its gradient.
When $\nx \leq 500$, the memory for all of the GP libraries is dominated by the memory of their dependencies and their own code rather than the matrices they are storing.
This baseline memory for GpFlash is smaller than for the other GP libraries in part since its code base is significantly smaller and it has fewer dependencies.
Table~\ref{Tab_speedup_gp_libs_mem_1t} shows how much more memory the GP libraries use compared to GpFlash.
For the case with $\nx = 10^4$ and $\ndim = 10$, where the memory usage is dominated by the data storage rather than dependencies, GpFlash uses 9.6 to 35 times less memory on Trillium and Mac.

For all of the GP libraries the memory grows as $\calO(\nx^2)$ for $\nx > 500$.
SMT and scikit-learn are the two GP libraries whose memory depends on $\ndim$.
SMT stores the componentwise distances between all pairs of points, which requires $\nx(\nx-1)\ndim/2$ doubles, while scikit-learn stores all $\ndim$ derivative matrices.

\begin{table}[t]
\centering\footnotesize
\setlength{\tabcolsep}{4pt}
\adjustbox{max width=\linewidth}{%
\begin{tabular}{l|ccc}
\hline
 & $n_x = 10^{2}$ & $n_x = 10^{3}$ & $n_x = 10^{4}$ \\
\hline
SMT & \green{55.9$\times$} / \green{52.1$\times$} & \green{41.5$\times$} / \green{32.6$\times$} & \green{24.6$\times$} / \green{14.5$\times$} \\
GPflow & \green{57.7$\times$} / \green{77.2$\times$} & \green{43.4$\times$} / \green{34.7$\times$} & \green{10.2$\times$} / \green{9.6$\times$} \\
scikit-learn & \green{11.2$\times$} / \green{21.8$\times$} & \green{21.5$\times$} / \green{24.3$\times$} & \green{34.6$\times$} / \green{14.7$\times$} \\
GPyTorch & \green{49.3$\times$} / \green{38.8$\times$} & \green{38.6$\times$} / \green{20.2$\times$} & \green{13.3$\times$} / \green{12.9$\times$} \\
GPy & \green{12.0$\times$} / \green{20.4$\times$} & \green{15.4$\times$} / \green{15.4$\times$} & \green{16.7$\times$} / \green{14.9$\times$} \\
\hline
Min ratio & \green{11.2$\times$} / \green{20.4$\times$} & \green{15.4$\times$} / \green{15.4$\times$} & \green{10.2$\times$} / \green{9.6$\times$} \\
Max ratio & \green{57.7$\times$} / \green{77.2$\times$} & \green{43.4$\times$} / \green{34.7$\times$} & \green{34.6$\times$} / \green{14.9$\times$} \\
\hline
\end{tabular}%
}

\caption{The table entries indicate how much more memory each GP library uses compared to GpFlash on Trillium / Mac.
The median memory was computed from 5 runs for problems run with $\ndim = 10$ and a single thread.
\green{Green} $\geq 2\times$.}
\label{Tab_speedup_gp_libs_mem_1t}
\end{table}

\section{Posterior evaluations} \label{Sec_PosteriorEval}

\subsection{Study 4: Evaluating the posterior mean and its gradient} \label{Sec_PosteriorEval__MeanAndGradEval}

Different methods of evaluating the mean of the posterior and its gradient with the use of a stationary kernel are presented in this section.
The following are precomputed for all cases in this subsection: $\Lmat$ from \eqref{Eq_Lmat}, $\vvecres$ from \eqref{Eq_vvecres}, and $\Xmatdot$ from \eqref{Eq_Xmatdot_point} along with its row norms.
The GP library scikit-learn is not considered since it does not provide an analytic gradient of the posterior, while the methods used by GPyTorch and GPflow are not represented since they use automatic differentiation, which is not supported in Armadillo.

\begin{description}[style=unboxed, leftmargin=0pt, labelindent=0pt, itemsep=2pt]
	\item[Case 1:] GPy and SMT use the polarization identity from Section~\ref{Sec_EfficientKernelEval__BLAS} to calculate $\kvec$ and \eqref{Eq_kvec_grad_xvec} to form the full Jacobian.
	\item[Case 2:] Similar to Case 1 but the evaluation of $\kvec$ and the Jacobian is done with \eqref{Eq_kern_efficient} and \eqref{Eq_kern_der_r_ell_efficient}, respectively.
	This reduces the number of exponential evaluations from $\nx (\ndim + 1)$ to $\nx$.
	\item[Case 3:] In GpFlash the gradient of the posterior mean is calculated with \eqref{Eq_Post_mean_grad_efficient}.
	The matrix $\Hmatr$ from \eqref{Eq_Hmat_r} is never formed since its calculation is fused with its multiplication to $\qvec_\mu$ from \eqref{Eq_qvec_mu}.
\end{description}

\begin{figure}[t]
\centering
\includegraphics[width=\textwidth]{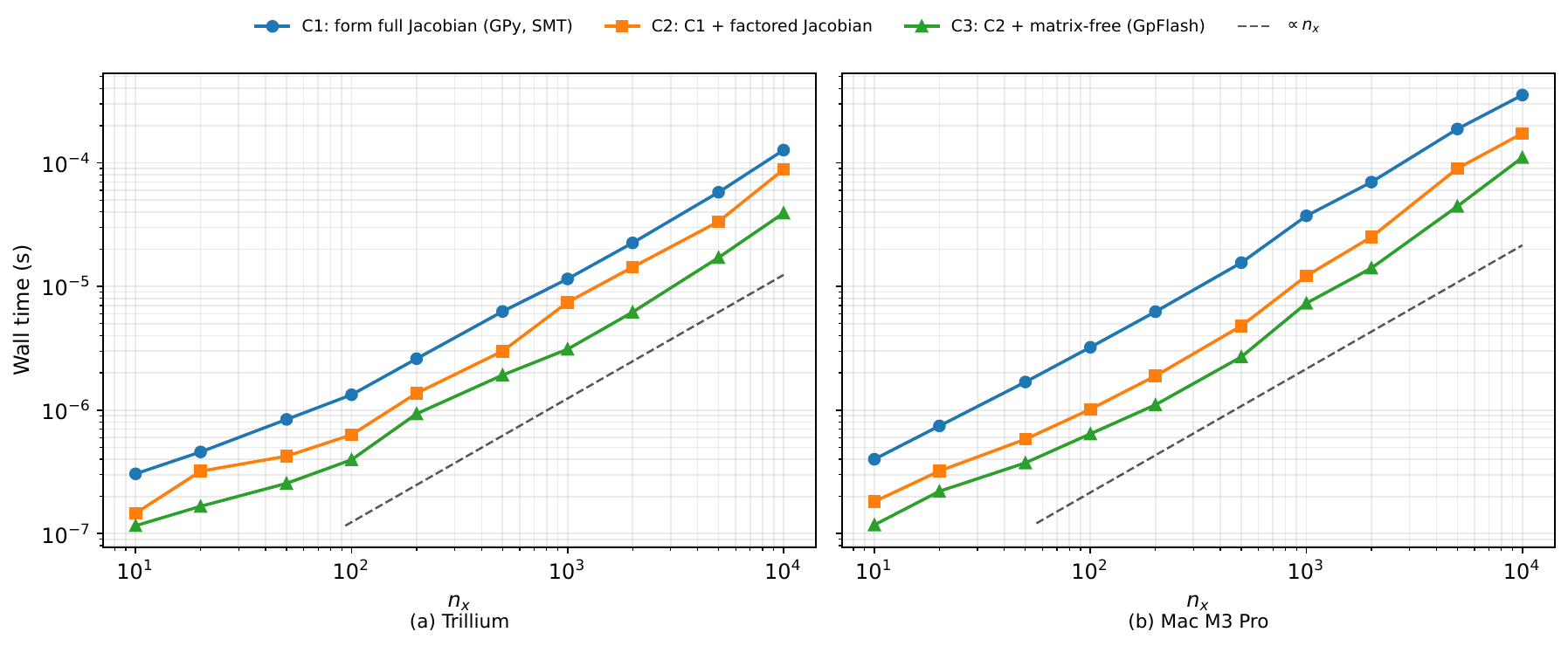}
\caption{Median wall time of 20 runs to evaluate the posterior mean and its gradient at one point with $\ndim = 10$ and 4 threads for the Mat\'ern $\frac{5}{2}$ kernel.}
\label{Fig_StudyPost_MeanGradient_10d_4threads}
\end{figure}

The wall time for the cases to evaluate the mean of the posterior and its gradient can be seen in Figure~\ref{Fig_StudyPost_MeanGradient_10d_4threads}, and they are compared in Table~\ref{Tab_study_post_grad_4t}.
Case 3, which is used by GpFlash, is 3.2 to 5.1 times faster than Case 1.

\begin{table}[t]
\centering\small
\adjustbox{max width=\linewidth}{%
\begin{tabular}{l|ccc}
\hline
 & $n_x = 10^{2}$ & $n_x = 10^{3}$ & $n_x = 10^{4}$ \\
\hline
C1: form full Jacobian (GPy, SMT) & baseline & baseline & baseline \\
C2: C1 + factored Jacobian & \green{2.1$\times$} / \green{3.2$\times$} & 1.5$\times$ / \green{3.1$\times$} & 1.4$\times$ / \green{2.0$\times$} \\
C3: C2 + matrix-free (GpFlash) & 1.6$\times$ / 1.6$\times$ & \green{2.4$\times$} / 1.7$\times$ & \green{2.3$\times$} / 1.6$\times$ \\
\hline
final vs.\ baseline & \green{3.4$\times$} / \green{5.0$\times$} & \green{3.7$\times$} / \green{5.1$\times$} & \green{3.2$\times$} / \green{3.2$\times$} \\
\hline
\end{tabular}
}

\caption{Study 4 (posterior mean and gradient evaluation): speedup each case provides relative to its predecessor from Figure~\ref{Fig_StudyPost_MeanGradient_10d_4threads} for Trillium / Mac.
\green{Green} $\geq 2\times$.}
\label{Tab_study_post_grad_4t}
\end{table}

\subsection{Cross-library comparison} \label{Sec_PosteriorEval__libComparison}

\begin{figure}[t]
\centering
\includegraphics[width=\textwidth]{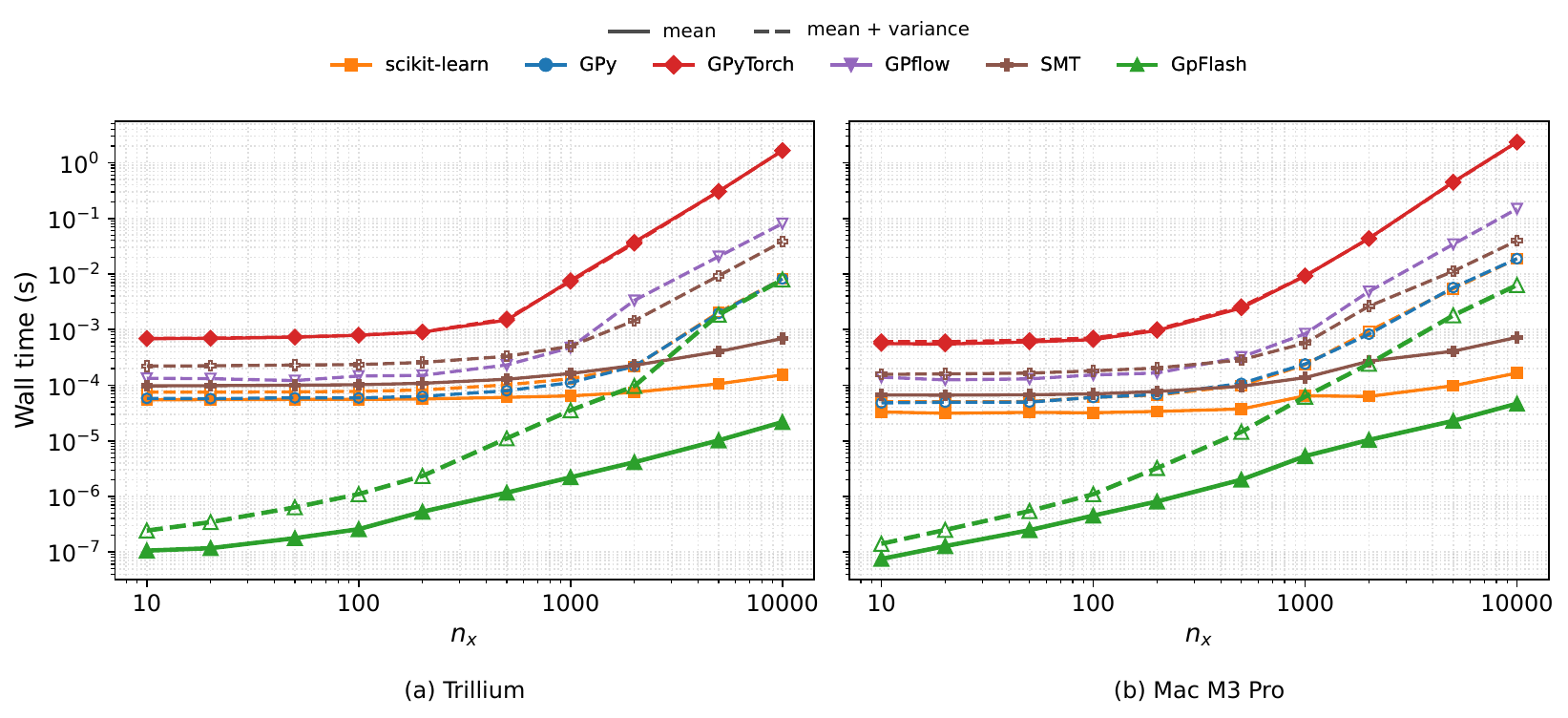}
\caption{Median wall time of 20 runs with 4 threads for a 10-dimensional problem with the Mat\'ern $\frac{5}{2}$ kernel to evaluate one point in the posterior.
The solid lines indicate the time to evaluate just the posterior mean while the dashed lines indicate the time to also evaluate the variance of the posterior.
Some libraries do not have a mean-only evaluation and thus have no solid line.}
\label{Fig_StudyPost_GpLibs_NoVar}
\end{figure}

The solid lines in Figure~\ref{Fig_StudyPost_GpLibs_NoVar} are the median wall time to evaluate the mean of the posterior at one point while the dashed lines are for the evaluation of the mean and variance of the posterior.
Several GP libraries always evaluate the mean and the variance together and thus only have a dashed line.

Table~\ref{Tab_speedup_gp_libs_post_novar_4t} indicates that GpFlash is at least 3.5 times faster than all of the GP libraries on both Trillium and Mac for the mean-only posterior evaluations.
For the mean and variance posterior evaluations the computational cost is dominated by the triangular solve for $\vveckhalf$ in \eqref{Eq_vveckhalf}.
At $\nx = 10^4$, the wall time of scikit-learn and GPy is within 4\% of that for GpFlash on Trillium, while GpFlash is 3 times faster than them on Mac.
This difference in performance on Mac is caused by the BLAS-3 method \texttt{DTRSM} running slower on Mac compared to the BLAS-2 vector solve \texttt{DTRSV}, which is used by GpFlash to evaluate a single point for the posterior.

\begin{figure}[t]
\centering
\includegraphics[width=\textwidth]{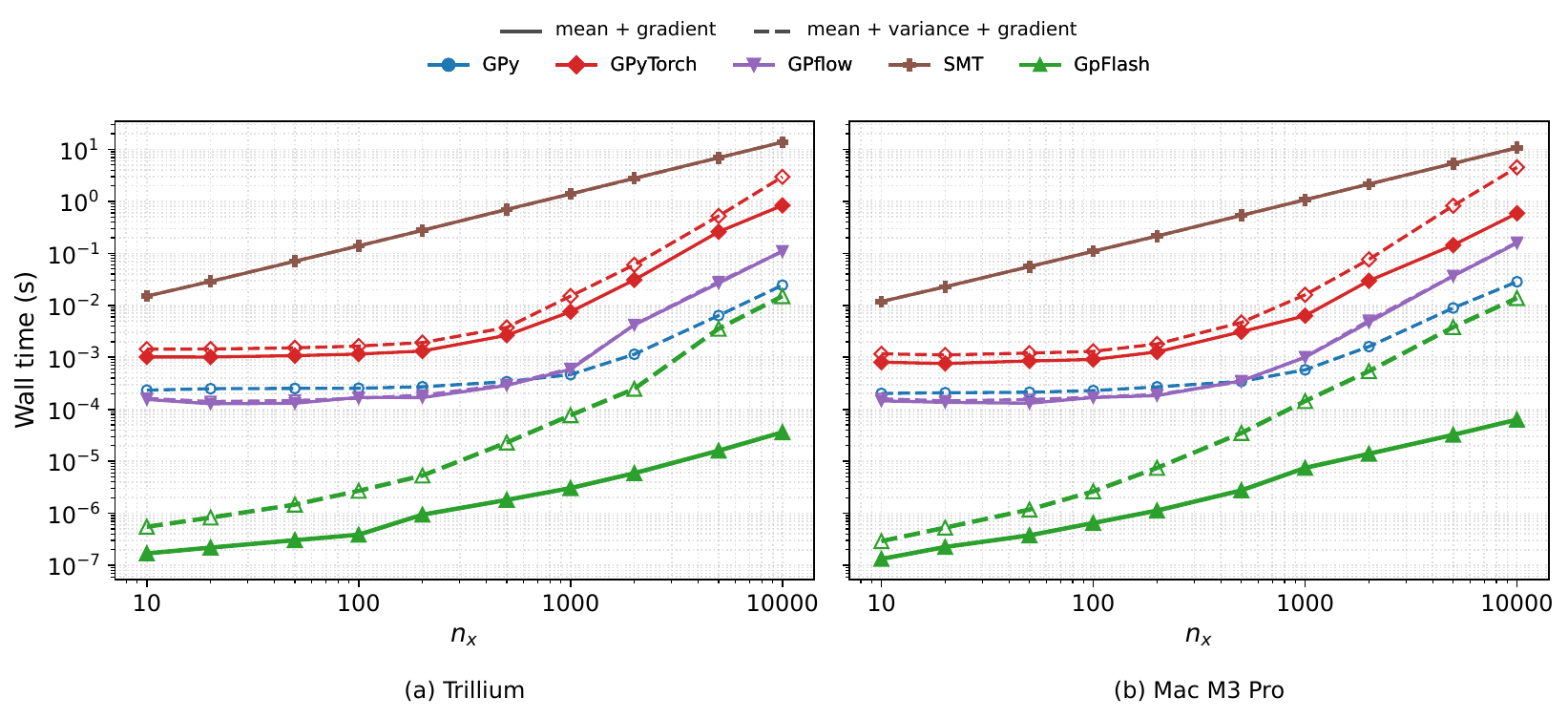}
\caption{The median wall time for the mean of the posterior and its gradient is shown with solid lines.
The dashed lines indicate the median wall time to evaluate the mean and variance of the posterior along with their gradients.
Twenty runs were performed using 4 threads and a data set with 10 dimensions for the Mat\'ern $\frac{5}{2}$ kernel.}
\label{Fig_StudyPost_GpLibs_Grad}
\end{figure}

Figure~\ref{Fig_StudyPost_GpLibs_Grad} along with Table~\ref{Tab_speedup_gp_libs_post_grad_4t} compares the wall time required for GpFlash and the other GP libraries to evaluate the mean of the posterior along with its gradient (solid lines), and the mean and variance of the posterior along with their gradients (dashed lines).
Only GpFlash and SMT offer the functionality to evaluate the mean of the posterior and its gradient without also evaluating the variance.
For the cases with $\nx = 10^4$, GpFlash is over 440 times faster than all of the other GP libraries for the evaluation of the mean of the posterior and its gradient.
When the variance of the posterior and its gradient are also evaluated, the speed advantage for GpFlash is reduced to 1.6 and 2.1 times on Trillium and Mac, respectively, since this requires expensive triangular solves.

\begin{table}[t]
\centering\footnotesize
\setlength{\tabcolsep}{4pt}
\adjustbox{max width=\linewidth}{%
\begin{tabular}{l|ccc|ccc}
\hline
 & \multicolumn{3}{c|}{Posterior mean} & \multicolumn{3}{c}{Posterior mean $+$ variance} \\
 & $n_x = 10^{2}$ & $n_x = 10^{3}$ & $n_x = 10^{4}$ & $n_x = 10^{2}$ & $n_x = 10^{3}$ & $n_x = 10^{4}$ \\
\hline
SMT & \green{395} / \green{154} & \green{73.4} / \green{25.8} & \green{31.8} / \green{15.5} & \green{214} / \green{165} & \green{14.2} / \green{9.2} & \green{4.8} / \green{6.3} \\
scikit-learn & \green{215} / \green{70.6} & \green{29.1} / \green{12.1} & \green{7.0} / \green{3.5} & \green{70.4} / \green{55.3} & \green{3.7} / \green{3.8} & 1.03 / \green{3.0} \\
GPy & \green{229} / \green{133}\textsuperscript{*} & \green{50.3} / \green{45.3}\textsuperscript{*} & \green{376} / \green{405}\textsuperscript{*} & \green{53.7} / \green{55.0} & \green{3.1} / \green{3.9} & 1.03 / \green{3.0} \\
GPyTorch & \green{3089} / \green{1564}\textsuperscript{*} & \green{3318} / \green{1738}\textsuperscript{*} & \green{$7.6{\times}10^{4}$} / \green{$5.0{\times}10^{4}$}\textsuperscript{*} & \green{724} / \green{645} & \green{207} / \green{148} & \green{208} / \green{371} \\
GPflow & \green{567} / \green{338}\textsuperscript{*} & \green{215} / \green{158}\textsuperscript{*} & \green{3708} / \green{3184}\textsuperscript{*} & \green{133} / \green{140} & \green{13.4} / \green{13.5} & \green{10.2} / \green{23.5} \\
\hline
Min speedup & \green{215} / \green{70.6} & \green{29.1} / \green{12.1} & \green{7.0} / \green{3.5} & \green{53.7} / \green{55.0} & \green{3.1} / \green{3.8} & 1.03 / \green{3.0} \\
Max speedup & \green{3089} / \green{1564} & \green{3318} / \green{1738} & \green{$7.6{\times}10^{4}$} / \green{$5.0{\times}10^{4}$} & \green{724} / \green{645} & \green{207} / \green{148} & \green{208} / \green{371} \\
\hline
\end{tabular}%
}

\caption{Speedup of GpFlash over each GP library on Trillium / Mac to evaluate one point in the posterior for the cases plotted in Figure~\ref{Fig_StudyPost_GpLibs_NoVar}.
An asterisk (\textsuperscript{*}) is used to note that GPy, GPflow, and GPyTorch do not have a mean-only interface.
\green{Green} $\geq 2\times$.}
\label{Tab_speedup_gp_libs_post_novar_4t}
\end{table}

\begin{table}[t]
\centering\footnotesize
\setlength{\tabcolsep}{4pt}
\adjustbox{max width=\linewidth}{%
\begin{tabular}{l|ccc|ccc}
\hline
 & \multicolumn{3}{c|}{Posterior mean $+$ gradient} & \multicolumn{3}{c}{Posterior mean $+$ variance $+$ gradient} \\
 & $n_x = 10^{2}$ & $n_x = 10^{3}$ & $n_x = 10^{4}$ & $n_x = 10^{2}$ & $n_x = 10^{3}$ & $n_x = 10^{4}$ \\
\hline
SMT & \green{$3.6{\times}10^{5}$} / \green{$1.7{\times}10^{5}$} & \green{$4.6{\times}10^{5}$} / \green{$1.4{\times}10^{5}$} & \green{$3.8{\times}10^{5}$} / \green{$1.7{\times}10^{5}$} & \green{$5.1{\times}10^{4}$} / \green{$4.1{\times}10^{4}$} & \green{$1.8{\times}10^{4}$} / \green{7467} & \green{921} / \green{770} \\
GPy & \green{657} / \green{352}\textsuperscript{*} & \green{152} / \green{76.9}\textsuperscript{*} & \green{674} / \green{449}\textsuperscript{*} & \green{94.0} / \green{86.7} & \green{6.1} / \green{4.0} & 1.6 / \green{2.1} \\
GPyTorch & \green{4251} / \green{2002}\textsuperscript{*} & \green{4929} / \green{2134}\textsuperscript{*} & \green{$8.2{\times}10^{4}$} / \green{$7.1{\times}10^{4}$}\textsuperscript{*} & \green{608} / \green{493} & \green{196} / \green{111} & \green{197} / \green{327} \\
GPflow & \green{422} / \green{262}\textsuperscript{*} & \green{199} / \green{135}\textsuperscript{*} & \green{3007} / \green{2579}\textsuperscript{*} & \green{60.4} / \green{64.6} & \green{7.9} / \green{7.0} & \green{7.2} / \green{11.8} \\
\hline
Min speedup & \green{422} / \green{262} & \green{152} / \green{76.9} & \green{674} / \green{449} & \green{60.4} / \green{64.6} & \green{6.1} / \green{4.0} & 1.6 / \green{2.1} \\
Max speedup & \green{$3.6{\times}10^{5}$} / \green{$1.7{\times}10^{5}$} & \green{$4.6{\times}10^{5}$} / \green{$1.4{\times}10^{5}$} & \green{$3.8{\times}10^{5}$} / \green{$1.7{\times}10^{5}$} & \green{$5.1{\times}10^{4}$} / \green{$4.1{\times}10^{4}$} & \green{$1.8{\times}10^{4}$} / \green{7467} & \green{921} / \green{770} \\
\hline
\end{tabular}%
}

\caption{Speedup of GpFlash over each GP library on Trillium / Mac for the data plotted in Figure~\ref{Fig_StudyPost_GpLibs_Grad}.
The library scikit-learn is omitted since it does not evaluate the gradient of the posterior analytically.
GPy, GPyTorch, and GPflow always evaluate the variance of the posterior along with the mean and this is indicated by an asterisk (\textsuperscript{*}).
\green{Green} $\geq 2\times$.}
\label{Tab_speedup_gp_libs_post_grad_4t}
\end{table}

\section{Conclusions} \label{Sec_Conclusions}

This paper presents GpFlash and demonstrates how to efficiently calculate the marginal log likelihood $\ln(L)$ and the posterior along with their gradients.
An efficient factorization is presented that reduces the number of exponential evaluations for stationary kernels from $\nx^2 (\ndim + 1)$ to $\nx^2/2$ for the evaluation of $\ln(L)$ with its gradient.
Applying this same factorization strategy to the posterior with gradient evaluations reduces the number of exponential evaluations from $\nx (\ndim + 1)$ to only $\nx$.

Studies compared different methods of calculating $\ln(L)$ and its gradient on the supercomputer Trillium and a MacBook Pro with an M3 Pro chip.
GpFlash was compared to the GP libraries scikit-learn, GPy, GPflow, GPyTorch, and SMT and was found to be faster than all of them for all cases considered.
For the case with $\nx=10^4$ training points and 10 dimensions, GpFlash was 2.6 to 54 times faster.
The computational speedup provided by GpFlash is even larger for smaller $\nx$ where the wall time is less dominated by BLAS calculations.

GpFlash uses a single $\nx \times \nx$ matrix $\Bmat$ for the log likelihood and gradient evaluations, holding two symmetric matrices at once in its lower and upper triangular entries, with the upper triangle laid out for efficient column-major access.
GpFlash was found to use 9.6 to 35 times less memory than the other GP libraries for cases with $\nx=10^4$ and $\ndim =10$.
The peak memory of GpFlash scales as $\calO(\nx^2 + \nx\ndim)$, while its wall time at large $\nx$ is nearly independent of $\ndim$ since the $\calO(\nx^3)$ operations to calculate $\Lmat$ and $\Sigma^{-1}$ dominate.

GpFlash is at least 440 times faster to evaluate the mean of the posterior along with its gradient than the other GP libraries for cases with $\ndim=10$ and $\nx = 10^4$.
When the variance is also evaluated, GpFlash remains the fastest but its net benefit is reduced since the cost is dominated by triangular solves, which have efficient BLAS methods for all GP libraries to use.

GpFlash can be found in the git repository \url{https://github.com/marchildon/gpflash}.

The work presented in this paper could be expanded to develop more efficient algorithms for use with GPUs.
Another opportunity is to develop more efficient algorithms for gradient-enhanced GPs, which have a training cost for exact evaluations that scales as $\calO(\nx^3 (\ndim + 1)^3)$.

\section*{Acknowledgements}

The author thanks Dr.\ Jose Arnal and Dr.\ Marco Ceze for their helpful comments in the review of this manuscript.
Computations were performed on the Trillium supercomputer at the SciNet HPC Consortium.
SciNet is funded by Innovation, Science and Economic Development Canada; the Digital Research Alliance of Canada; the Ontario Research Fund: Research Excellence; and the University of Toronto.

\section*{Declaration of generative AI and AI-assisted technologies in the writing process}

During the preparation of this work the author used Anthropic models in order to proofread the manuscript and provide suggestions for improvements.
After using this tool, the author reviewed and edited the content as needed and takes full responsibility for the content of the publication.


\appendix

\section{Vanishing term for the gradient of the marginal log likelihood} \label{SecAppendix_dbetaVanish}

It is shown here that the derivative of $\hpbeta$ does not need to be calculated since its term in the gradient of $\ln(L)$ in \eqref{Eq_grad_ln_lkd_SMT} is zero:
\begin{align*}
	\p{\hpbeta}{\hpalpha} \one^\top \Sigma^{-1} \left( \fvectil - \hpbeta \one \right)
		&= \p{\hpbeta}{\hpalpha} \left( \one^\top \Sigma^{-1} \fvectil - \hpbeta \one^\top \Sigma^{-1} \one \right) \\
		&= \p{\hpbeta}{\hpalpha} \left( \one^\top \Sigma^{-1} \fvectil - \left( \frac{ \one^\top \Sigma^{-1} \fvectil}{ \one^\top \Sigma^{-1} \one} \right) \one^\top \Sigma^{-1} \one \right) \\
		&= 0.
\end{align*}

\bibliography{MyLibrary}

\end{document}